\documentclass[sigconf]{acmart}

\usepackage{booktabs} 

\usepackage[english]{babel}
\usepackage{moresize}
\usepackage{threeparttable} 
\usepackage{amsmath}
\usepackage{algorithmic}
\usepackage{balance}
\usepackage{comment}
\usepackage{paralist}
\usepackage{bm}
\usepackage{bbding}
\usepackage{pgfplots}
\usetikzlibrary{pgfplots.dateplot}
\usepackage{xcolor}
\usepackage{flushend}
\usepackage[english]{babel}
\usepackage[latin1]{inputenc}
\usepackage{mathrsfs}
\usepackage{graphicx}

\usepackage{amssymb}
\usepackage{amsfonts}
\usepackage{url}
\usepackage{longtable}
\usepackage{rotating}
\usepackage{multirow}
\usepackage{mathrsfs}
\usepackage{subfigure}
\usepackage{enumitem}
\usepackage[linesnumbered,algoruled,boxed,lined]{algorithm2e}
\usepackage{adjustbox}
\usepackage{hyperref}
\usepackage{pgfplots}
\usetikzlibrary{pgfplots.dateplot}
\usepackage{filecontents}
\usepackage{tabularx}
\usepackage{threeparttable}
\definecolor{tblue}{RGB}{31,119,180}
\definecolor{torange}{RGB}{255,127,14}
\definecolor{tgreen}{RGB}{44,160,44}
\definecolor{tred}{RGB}{214,39,40}
\definecolor{tpurple}{RGB}{148,103,189}

\usepackage{colortbl}
\usepackage{xcolor}
\usepackage{array}
\usepackage{listings}
\usepackage{multirow}
\usepackage{soul} 
\usepackage{color} 

\newcommand{\ie}{\textit{i}.\textit{e}.}
\newcommand{\eg}{\textit{e}.\textit{g}.}

\def\model{HiGPT}

\AtBeginDocument{%
  }

\setcopyright{none}

\begin{document}
\fancyhead{}
\title{HiGPT: Heterogeneous Graph Language Model}

\author{Jiabin Tang$^1$, Yuhao Yang$^2$, Wei Wei$^2$, Lei Shi$^3$, \\ Long Xia$^3$, Dawei Yin$^3$ and Chao Huang$^{1,2*}$}
\thanks{$*$ Chao Huang is the Corresponding Author.}
\affiliation{$^1$Musketeers Foundation Institute of Data Science, \\ $^2$Department of Computer Science,  University of Hong Kong, $^3$Baidu Inc. \country{}\\}
\affiliation{\textbf{Project Page}: \href{https://HiGPT-HKU.github.io}{https://HiGPT-HKU.github.io}, \textbf{Github}: \href{https://github.com/HKUDS/HiGPT}{https://github.com/HKUDS/HiGPT} \country{}}

\renewcommand{\shortauthors}{Anonymous Author(s)}

\begin{abstract}
Heterogeneous graph learning aims to capture complex relationships and diverse relational semantics among entities in a heterogeneous graph to obtain meaningful representations for nodes and edges. Recent advancements in heterogeneous graph neural networks (HGNNs) have achieved state-of-the-art performance by considering relation heterogeneity and using specialized message functions and aggregation rules. However, existing frameworks for heterogeneous graph learning have limitations in generalizing across diverse heterogeneous graph datasets. Most of these frameworks follow the "pre-train" and "fine-tune" paradigm on the same dataset, which restricts their capacity to adapt to new and unseen data. This raises the question: ``Can we generalize heterogeneous graph models to be well-adapted to diverse downstream learning tasks with distribution shifts in both node token sets and relation type heterogeneity?'' To tackle those challenges, we propose \model, a general large graph model with \underline{H}eterogeneous graph \underline{i}nstruction-tuning paradigm. Our framework enables learning from arbitrary heterogeneous graphs without the need for any fine-tuning process from downstream datasets. To handle distribution shifts in heterogeneity, we introduce an in-context heterogeneous graph tokenizer that captures semantic relationships in different heterogeneous graphs, facilitating model adaptation. We incorporate a large corpus of heterogeneity-aware graph instructions into our \model, enabling the model to effectively comprehend complex relation heterogeneity and distinguish between various types of graph tokens. Furthermore, we introduce the Mixture-of-Thought (MoT) instruction augmentation paradigm to mitigate data scarcity by generating diverse and informative instructions. Through comprehensive evaluations conducted in various settings, our proposed framework demonstrates exceptional performance in terms of generalization performance, surpassing current leading benchmarks.
\end{abstract}



\maketitle

\section{Introduction}
\label{sec:intro}

Heterogeneous graphs have garnered extensive popularity and adoption in various domains, including recommendation systems~\cite{fan2019metapath}, knowledge graphs~\cite{wang2021relational}, social network analysis~\cite{el2022twhin}, and biological networks~\cite{ma2023single}. These graphs encompass entities of diverse types that engage in a multitude of interactions, enabling a comprehensive representation of complex systems~\cite{dong2017metapath2vec}. The focus of heterogeneous graph learning is to derive meaningful representations for the nodes and edges within such graphs~\cite{zhao2021heterogeneous,lv2021we,jin2021heterogeneous}. These representations aim to capture the intricate relationships and diverse relational semantics that exist within the graph, facilitating a deeper understanding of the underlying structural heterogeneity.

In recent years, there has been a growing recognition of the significant potential of heterogeneous graph neural networks (HGNNs) in capturing the intricate and diverse information that resides within heterogeneous graph structures \cite{wang2022survey,yang2020heterogeneous}. HGNNs leverage the expressive capabilities of high-order message passing techniques, enabling them to effectively model the complex relationships, diverse entity types, and heterogeneous semantics present in these graphs. By aggregating and propagating information across various node and edge types, HGNNs facilitate a deeper understanding and analysis of the intricate inter-dependencies that exist within heterogeneous graph structures. Some notable examples of HGNNs include: i) Metapath-based GNNs such as HAN \cite{wang2019heterogeneous} and MAGNN \cite{fan2019metapath}; ii) Transformer-enhanced GNNs like HGT \cite{HGT}. Furthermore, heterogeneous graph self-supervised learning, including contrastive methods (\eg, DMGI \cite{park2020unsupervised} and HeCo \cite{wang2021self}), and generative (\eg, HGMAE \cite{tian2023heterogeneous}) methods, showcase effectiveness in alleviating data scarcity in real-world heterogeneous graph data.

Despite the demonstrated effectiveness of current frameworks for heterogeneous graph learning, they possess limitations when it comes to generalizing across diverse heterogeneous graph datasets. These frameworks commonly adopt the "pre-train" and "fine-tune" paradigm, where they are initially trained on a specific dataset and subsequently fine-tuned on the same dataset~\cite{HGT,wang2021self,tian2023heterogeneous}. However, this approach presents challenges in adapting and achieving optimal performance on new and unseen data. The heavy reliance on the characteristics and patterns of the original training dataset hinders their ability to effectively handle the intricacies and complexities inherent in different heterogeneous graph datasets. As a result, these frameworks may encounter difficulties in effectively addressing the diverse nuances and variations present in various heterogeneous graph datasets, particularly when applied to downstream tasks.

This research aims to push the boundaries of heterogeneous graph models by addressing a fundamental question: "Can we develop highly adaptable and versatile heterogeneous graph models capable of effectively addressing diverse downstream learning tasks, even when faced with distribution shifts in node token sets and relation type heterogeneity?" To tackle this challenge, we introduce \model\ as a novel and general solution. Our model is specifically designed to overcome key challenges associated with generalization across various downstream heterogeneous graph learning tasks.\\\vspace{-0.12in}

\noindent \textbf{C1. Relation Type Heterogeneity Shift}. One of the primary challenges we focus on in this research is the shift in relation type heterogeneity within various heterogeneous graph structures. In such graphs, entities are connected by various types of relations, and these relation types can differ significantly across diverse datasets. To illustrate this, let's consider two examples. In a recommendation system, a heterogeneous graph may involve node-wise relationships between users and items. The relation types in this scenario could include "click," "favorite," "review," and "purchase." On the other hand, in an academic graph, the relations could involve "paper-paper," "author-paper," and "paper-venue." These examples demonstrate how different heterogeneous graphs can exhibit diverse relation heterogeneity with distinct semantics across domains. \\\vspace{-0.12in}

\noindent \textbf{Solution: In-Context Heterogeneous Graph Tokenizer}. To achieve adaptability in a wide range of heterogeneous graph scenarios with varying node and edge types, we introduce the in-context heterogeneous graph tokenizer. This tokenizer captures the diverse semantic relationships found in different heterogeneous graphs, providing a unified approach. It comprises two essential components: the in-context parameterized heterogeneity projector, which utilizes language to encode distinct node and edge types, facilitating effective model adaptation, and the parameter allocator, which dynamically assigns tailored parameters to the tokenizer. To optimize performance and integrate the tokenizer seamlessly into the \model\ framework, we employ pre-training with a lightweight text-graph contrastive alignment paradigm. This pre-training process directly incorporates the tokenizer into \model, enhancing its capabilities and ensuring smooth functionality within the overall model architecture, including integration with the language model. \\\vspace{-0.12in}

\noindent \textbf{C2. Complex Heterogeneous Graph Structures}. The primary focus of this study is to tackle the challenge of integrating large language models (LLMs) into heterogeneous graph learning, with the goal of enhancing model generalization. Our specific objective is to develop a graph-oriented language model that excels in comprehending the intricate structural information inherent in complex heterogeneous graph structures. In doing so, we strive to equip the graph model with the ability to not only recognize the heterogeneity of relations among different types of nodes, but also capture the distinct characteristics of entities belonging to the same type. \\\vspace{-0.12in}

\noindent \textbf{Solution: Heterogeneous Graph Instruction-Tuning}. We introduce a novel heterogeneous graph instruction-tuning framework that integrates inter-type and intra-type token matching tasks to fine-tune large language models (LLMs). Our framework specifically targets the enhancement of LLMs' understanding of both heterogeneous relation awareness and homogeneous relation awareness. By utilizing these tasks, our aim is to bolster the LLMs' capabilities in the following areas: (i) distinguishing between different types of graph tokens, (ii) comprehending intricate relationships within heterogeneous graphs, (iii) preserving the distinctive attributes of entities within homogeneous graphs, and (iv) effectively harnessing diverse graph instructions during the training process. \\\vspace{-0.12in}

\noindent \textbf{C3. Data Scarcity for Model Fine-Tuning}. In practical scenarios involving heterogeneous graph learning, one of the key challenges is the limited availability of data~\cite{hwang2020self,yang2022self}. This scarcity of data poses a significant obstacle when fine-tuning models for real-world applications. For instance, when utilizing heterogeneous graphs to model cold-start users or items in recommendation systems, the sparse nature of user interaction data restricts the availability of supervised signals~\cite{xuan2023knowledge,wei2022contrastive}. This data scarcity hampers the effectiveness of task-specific model fine-tuning and necessitates the development of novel techniques to overcome this challenge. \\\vspace{-0.12in}

\noindent \textbf{Solution: Mixture-of-Thought Augmentation}. Our approach introduces a novel mechanism for augmenting graph instructions, emphasizing the use of Mixture-of-Thought (MoT) combined with various prompting techniques. This integration enables us to generate a diverse and comprehensive set of informative task-specific instructions. By seamlessly incorporating these augmented graph instructions into our framework, we anticipate that our model enhancement will effectively address the challenge of data sparsity. \\\vspace{-0.12in}

\noindent \textbf{Experiments}. To assess the efficacy of our proposed approach, we perform a comprehensive set of experiments to showcase the impressive generalization capabilities of our heterogeneous graph language model across diverse settings. We thoroughly investigate the design rationales, effectiveness, and efficiency of our model. 
\section{Preliminaries}
\label{sec:model}

\noindent\textbf{Heterogeneous Graph}.
A heterogeneous graph is a graph denoted as $\mathcal{G}(\mathcal{V}, \mathcal{E}, \mathbf{A}, \mathcal{T}, \mathcal{R}, \mathbf{X})$. It consists of nodes represented by $\mathcal{V}$, edges represented by $\mathcal{E}$, and an adjacency matrix $\mathbf{A}$ that captures the relationships between nodes. The sets $\mathcal{T}$ and $\mathcal{R}$ signify the types of nodes and edges, respectively. Additionally, the feature matrix $\mathbf{X} = \{X_{T_i} \in \mathbb{R}^{|\mathcal{V}_{T_i}|\times d_{T_i}}\}$ contains attributes associated with each node. Here, $T_i$ refers to a specific node type, while $d_{T_i}$ represents the dimensionality of the corresponding node attributes. \\\vspace{-0.12in}

\noindent\textbf{Meta Relation}.
In a heterogeneous graph, a meta relation is a representation of the relationship between different types of nodes connected by an edge. Specifically, for an edge $e$ that links a source node $u$ of type $T_{i}$ to a target node $v$ of type $T_{j}$, the meta relation of $e = (u, v)$ is denoted as $<\tau(u), \rho(e), \tau(v)>$. Here, $\tau(u)$ and $\tau(v)$ represent the node types of $u$ and $v$ respectively, drawn from the set of node types $\mathcal{T}$, while $\rho(e)$ denotes the relation type of the edge $e$, which is selected from the set of relation types $\mathcal{R}$. This meta relation provides a concise representation of the heterogeneous connections in the graph, capturing the types of nodes involved. \\\vspace{-0.12in}


\noindent\textbf{Heterogeneous Graph Neural Networks (HGNNs)}. 
In the context of a heterogeneous graph $\mathcal{G}$, Heterogeneous Graph Neural Networks (HGNNs) employ message passing and aggregation techniques to incorporate neighbor information based on different node and edge categories. This enables the modeling of heterogeneous structural semantic relationships, as expressed below:
\begin{align}
    h_{v}^{(l)} = \underset{\forall u \in \mathcal{N}(v), \forall e \in \mathcal{E}(u, v)} {\textbf{Aggregate}} \left(\textbf{Propagate}\left(h_{u}^{(l-1)}; h_{v}^{(l - 1)}, e\right)\right) \label{eq:hgnn}
\end{align}
Here, $\mathcal{N}(v)$ represents all the source nodes connected to node $v$, and $\mathcal{E}(u, v)$ denotes the set of edges connecting node $u$ and node $v$. In most HGNNs, the parameters of the $\textbf{Propagate}\left(\cdot\right)$ and $\textbf{Aggregate}\left(\cdot\right)$ functions depend on the types of nodes $u$ and $v$, as well as the edge $e$. However, this implies that HGNNs are limited to modeling the specific heterogeneous graph they were trained on and cannot be effectively applied to new heterogeneous graphs with different node and edge types. This limitation greatly hampers the generalization ability of HGNNs in capturing the diverse semantic relationships across various heterogeneous graphs.

\section{Methodology}
\label{sec:solution}

\begin{figure*}[h]
    \centering
    \includegraphics{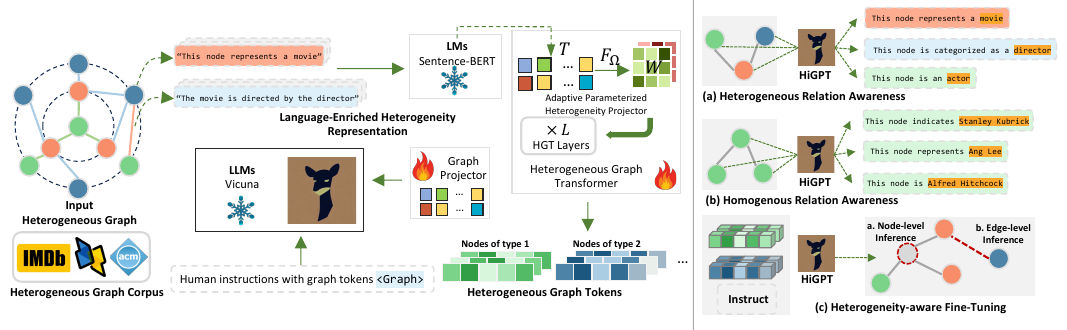}
    \vspace{-0.3in}
    \caption{The overall architecture of our \model.}
    \vspace{-0.15in}
    \label{fig:overall}
\end{figure*}



\subsection{In-Context Heterogeneous Graph Tokenizer}
To make our \model\ adaptable to a wide range of heterogeneous graph scenarios with varying node and edge types, we propose an in-context heterogeneous graph tokenizer. This method captures the diverse semantic relationships present in different heterogeneous graphs, ensuring a unified approach. It comprises two essential components: the in-context parameterized heterogeneity projector and the parameter allocator. The adaptive in-context projector utilizes language to encode the distinct node and edge types within the heterogeneous graphs, facilitating model adaptation. 

Meanwhile, the parameter allocator dynamically assigns parameters tailored specifically for the tokenizer. To optimize the tokenizer's performance and seamlessly integrate it within the \model\ framework, we conduct pre-training using a simplified text-graph contrastive learning framework. This pre-training process directly incorporates the tokenizer into the \model\ framework and effectively integrates it with the Large Language Model (LLM). This approach enhances the tokenizer's capabilities and ensures its smooth functioning within the overall model architecture.

\vspace{-0.1in}
\subsubsection{\bf Graph Tokenization with Meta Projector} Given a heterogeneous graph $\mathcal{G}$ with a feature matrix $\mathbf{X} = \{X_{T_i} \in \mathbb{R}^{|\mathcal{V}_{T_i}|\times d_{T_i}}, T_i \in \mathcal{T}\}$ and an adjacency matrix $\mathbf{A}$, the goal of the heterogeneous graph tokenizer is to encode the hidden representations of the heterogeneous graph, denoted as $\mathbf{H} = \{H_{T_i} \in \mathbb{R}^{|\mathcal{V}_{T_i}|\times f_{T_i}}, T_i \in \mathcal{T}\}$. This is achieved through the function $\mathbf{H} = \text{HG-Tokenizer}(\mathbf{X}, \mathbf{A})$, where $f_{T_i}$ represents the hidden dimension of node type $T_i$. The $\text{HG-Tokenizer}(\cdot)$ can be implemented using various backbone HGNN architectures, such as HetGNN~\cite{hetgnn}, HAN~\cite{han}, or HGT~\cite{HGT}.

However, the generalization capability of these heterogeneous GNNs is constrained by their inherent design, which includes pre-defined parameter learning tailored to specific heterogeneous graphs. As a result, the trained HGNNs cannot be readily applied to other unseen heterogeneous graphs, which goes against the objective of achieving unified encoding with the HG-Tokenizer. To illustrate, let's consider HGT as an example. In HGT, the calculation of $h_{v}^{(l)}$ involves utilizing functions such as $\mathbf{Attention}(\cdot)$ and $\mathbf{Message}(\cdot)$ to process information from the source nodes:
\begin{align}
    \widetilde{h}_{v}^{(l)} &= \underset{\forall u \in \mathcal{N}(v)}{\oplus} \left(\mathbf{Attention}\left(u,e,v\right)	\cdot \mathbf{Message}\left(u,e,v\right)\right) \nonumber \\ 
    h_{v}^{(l)} &= \mathcal{F}_{\Theta_1}^{\tau(v)} \left(\sigma\left(\widetilde{h}_{v}^{(l)}\right)\right) + h_{v}^{(l - 1)} \nonumber \\
    &= \mathbf{W}^{\tau(v)}_1\cdot \left(\sigma\left(\widetilde{h}_{v}^{(l)}\right)\right) + \mathbf{b}^{\tau(v)}_1 + h_{v}^{(l - 1)}
\end{align}
The notation $\mathcal{F}_{\Theta_1}^{\tau(v)}\left(\cdot\right)$ represents a fully-connected layer with parameters $\Theta_1 = \{\mathbf{W}^{\tau(v)}, \mathbf{b}^{\tau(v)}\}$. Here, $\tau(v)$ denotes the node type of $v$, and $\sigma(\cdot)$ represents the activation function. The specific formulation of the $\mathbf{Attention}(\cdot)$ and $\mathbf{Message}(\cdot)$ functions, with $h$ heads:
\begin{align}
&\mathbf{Attention}\left(u,e,v\right) \nonumber \\ &= \underset{\forall u \in \mathcal{N}(v)}{\text{Softmax}}\left(\underset{i\in [1, h]}{\parallel} \mathcal{F}_{\Theta_2}^{\tau(u)}\left(h_{u}^{(l - 1)}\right)\mathbf{W}^{\rho(e)}_1   \mathcal{F}_{\Theta_3}^{\tau(v)}\left(h_{v}^{(l - 1)}\right)\right) \nonumber \\
&\mathbf{Message}\left(u,e,v\right) = \underset{i\in [1, h]}{\parallel} \mathcal{F}_{\Theta_4}^{\tau(u)}\left(h_{u}^{(l - 1)}\right)\mathbf{W}^{\rho(e)}_2 
\end{align}

\noindent \textbf{Adaptive Parameterized Heterogeneity Projector}.
To make our \model\ adaptable to a wide range of heterogeneous graphs with varying graph heterogeneity settings, and to eliminate the requirement of pre-defining the number of type-specific projections in advance, we propose the design of a type-aware parameterized projector. This projector dynamically and automatically encodes the relation heterogeneity into latent representations. More specifically, the type-aware projectors with the parameters $\mathcal{F}_{\Theta_i}^{\tau(v)}$ and $\mathbf{W}^{\rho(e)}_i$ are generated automatically according to the following procedure:
\begin{align}
\Theta_i = \{\mathbf{W}^{\tau(v)}_i;~ \mathbf{b}^{\tau(v)}_i \}&= \mathcal{F}_{\Omega}\left(\mathbf{T}^{\tau(v)}\right);~~~\mathbf{W}^{\rho(e)}_i &= \mathcal{F}_{\Omega}\left(\mathbf{T}^{\rho(e)}\right) 
\end{align}
$\mathcal{F}_{\Omega}$ is a fully-connected layer with parameters $\Omega$, while $\mathbf{T}^{\tau(v)}$ and $\mathbf{T}^{\rho(e)}$ are the features associated with node type $\tau(v)$ and edge type $\rho(e)$, respectively. It is important to note that the example provided showcases the usage of the in-context parameterized heterogeneity projector within the heterogeneous graph transformer framework. However, our \model\ is designed to be versatile and adaptable, allowing for the integration of diverse heterogeneous GNNs. \\\vspace{-0.12in}

\noindent\textbf{Language-Enriched Heterogeneity Representation}.
We leverage natural language as a means to generate universal heterogeneity representations for nodes and edges based on their respective types. For instance, in the heterogeneous IMDB dataset, we can describe a "movie" node as "This node represents a movie" using natural language. Similarly, the edge ("movie", "to", "director") can be expressed as "The movie is directed by the director". To encode these natural language descriptions of nodes and edges, we employ a pre-trained language model such as Sentence-BERT~\cite{sent_bert} to obtain type representations.  To ensure distinguishability and diversity among different types, we utilize multiple languages to describe the same type. The encoded representations from the pre-trained language models are averaged to derive the final representation. This process can be defined as follows:
\begin{align}
\mathbf{T}^{\tau(v)} &= \text{Mean-Pooling}\left(\text{Sentence-BERT}\left( \mathbf{S}^{\tau(v)}\right)\right) \nonumber \\ 
\mathbf{T}^{\rho(e)} &= \text{Mean-Pooling}\left(\text{Sentence-BERT}\left( \mathbf{S}^{\rho(e)}\right)\right)
\end{align}
$\mathbf{S}^{\tau(v)}$ and $\mathbf{S}^{\rho(e)}$ represent sets of descriptions for node type $\tau(v)$ and edge type $\rho(e)$, respectively. For instance, consider the example of the edge ("movie", "to", "director"). One possible description is:
\begin{align}
&\mathbf{S}^{(\texttt{"movie"}, \texttt{"to"}, \texttt{"director"})} = \{ \nonumber \\
&\texttt{"The movie is directed by the director"}, \nonumber \\
&\texttt{"The film features direction by the director"}, \cdots\}
\end{align}
For comprehensive descriptions featuring text-enriched heterogeneity representations of various datasets, please consult the appendix.

\subsubsection{\bf Lightweight Text-Graph Contrastive Alignment}
Building upon recent advancements in aligning cross-modality semantics~\cite{CLIP, G2P2}, we draw inspiration to employ a text-graph contrastive alignment paradigm for pre-training the proposed heterogeneous graph tokenizer. This approach aims to align the modeling capabilities of language and heterogeneous structures, enabling better collaboration between the tokenizer and the language models. To begin, we consider raw textual contents represented as $\mathbf{C} = { c_i \in \mathbb{R}^{l_i\times d}, 1\leq i\leq N}$, where $N$ denotes the total number of heterogeneous graph nodes $\mathbf{X} = {X_{T_i} \in \mathbb{R}^{|\mathcal{V}_{T_i}|\times d_{T_i}}}$. Here, $l_i$ represents the length of the textual content associated with the $i$-th node. In our approach, we adopt a lightweight text-graph contrastive alignment paradigm formally presented as follows:
\begin{align}
\hat{\mathbf{H}} &= \text{norm}\left(\text{HG-Tokenizer}\left(\mathbf{X}\right)\right),
\hat{\mathbf{T}} = \text{norm}\left(\text{LM-Tokenizer}\left(\mathbf{C}\right)\right) \nonumber \\
\mathcal{L} &= \frac{1}{2}\left(\text{CE}(\Lambda, \mathbf{y}) + \text{CE}(\Lambda^{\top}, \mathbf{y})\right), \Lambda = (\hat{\mathbf{H}} \hat{\mathbf{T}}^{\top}) \cdot \exp (\tau)
\end{align}
We use the contrastive label $\mathbf{y} = (0, 1, \cdots, n-1)^{\top}$ and the Cross-Entropy function $\text{CE}(\cdot)$. Our implementation employs the multi-layer vanilla transformer for $\text{LM-Tokenizer}(\cdot)$.

\begin{table*}
  \centering
  \small
  \caption{Prompts for the three tasks of heterogeneous graph instruction-tuning.}
  \vspace{-0.15in}
  \label{tab:prompt}
  \resizebox{0.95\linewidth}{!}{
  \begin{tabular}{p{3.5cm}p{11cm}p{3cm}}
    \toprule
    HeteroGraph & Human Question & \model~Response \\
    \midrule
    \multicolumn{3}{c}{(a) Heterogeneous Instruction Pre-training} \\
    \cmidrule(rl){1-3}
    central\_nodes: ("movie": [1, ..., n]), num\_neighbors:[10, 10]
    & 
    Given a heterogeneous graph about \textcolor{teal}{movies}, there are \textcolor{teal}{3 types of nodes: <DESC>}. By performing random sampling, a \textbf{heterogeneous subgraph} is obtained. Separately \textcolor{violet}{nodes of different types} are: 1. \textcolor{teal}{<graph>}, 2. \textcolor{orange}{<graph>}... Please sequentially \textbf{provide the types} for the node sequences.
    & 
    Based on graph tokens, types of the graph tokens should be \textcolor{teal}{1. movie, 2. ...}.
    \\
    \midrule
    \multicolumn{3}{c}{(b) Homogenous Instruction Pre-training} \\
    \cmidrule(rl){1-3}
    central\_nodes: ("paper": [1, ..., n]), num\_neighbors: [10, 10]
    &
    Given a heterogeneous graph about \textcolor{brown}{papers}, there are \textcolor{brown}{4 types of nodes: <DESC>}. ..., a \textbf{heterogeneous subgraph} is obtained. The \textcolor{brown}{nodes for "paper"} are: \textcolor{brown}{<graph>}. Also, a list of textual descriptions for the papers are: \textcolor{brown}{<DESC>}. Please \textbf{reorder the text list based on the order of graph tokens}.
    & 
    The matching of graph tokens and papers should be: \textcolor{brown}{<ANSWER>}.
    \\
    \midrule
    \multicolumn{3}{c}{(c) Heterogeneous Supervised Fine-Tuning} \\
    \cmidrule(rl){1-3}
    central\_nodes: ("movie": [i]), num\_neighbors: [10, 10]
    &
    Given a heterogeneous graph about \textcolor{teal}{movies}, there are \textcolor{teal}{3 types of nodes: <DESC>}. ..., a \textbf{heterogeneous subgraph} is obtained. There are \textcolor{violet}{nodes of different types}: \textcolor{teal}{"movie" nodes}: \textcolor{teal}{<graph>}, \textcolor{teal}{<DESC>} where the 0-th node is the central node. \textcolor{orange}{"actor" nodes}: \textcolor{orange}{<graph>}; \textcolor{pink}{"director" nodes}: \textcolor{pink}{<graph>}. \textbf{Which of the following classes does this \textcolor{teal}{movie} belong to}: action, comedy, drama?
    &
    Based on the given information, the likely category for \textcolor{teal}{movie} is \textcolor{teal}{Action}.
    \\
    \bottomrule
  \end{tabular}
  }
\end{table*}

\subsection{Heterogeneous Graph Instruction Tuning}
The objective of \model\ is to empower language models to directly generate predictions for downstream tasks with the unseen heterogeneous graph and corresponding instructions. The natural language instruction is first encoded by a tokenizer into text embeddings, denoted as $\mathbf{X}_{\mathcal{I}} = \text{LM-tokenizer}(\texttt{instruction})$. To align the dimensions, we employ a projector that maps graph tokens to the same dimension as the text embeddings, given by $\mathbf{X}_{\mathcal{G}} = f_{\mathbf{P}}(\mathbf{H})$, which can be as simple as a linear layer. For a sequence of length $L$, we determine the probability of generating the target output $\mathbf{X}_{\mathcal{O}}$:
\begin{align}
p(\mathbf{X}_{\mathcal{O}}|\mathbf{X}_{\mathcal{G}}, \mathbf{X}_{\mathcal{I}}) = \prod_{i=1}^{L}p_{\Phi}(x_i|\mathbf{X}_{\mathcal{G}}, \mathbf{X}_{\mathcal{I}, <i} , \mathbf{X}_{\mathcal{O}, <i})
\end{align}
where $\Phi$ represents the learnable parameters within \model.

\subsubsection{\bf Instruction Tuning with Heterogeneous Graph Corpus}
To enable the Language Model (LLM) to effectively differentiate between different types of input heterogeneous graph tokens and the specific nodes within each type, based on natural language instructions, we propose instruction pre-training using a large "corpus" consisting of heterogeneous graph-instruction pairs. This approach equips the fine-tuned \model\ with a comprehensive understanding of both homogeneous and heterogeneous graph structures.

\begin{itemize}[leftmargin=*]

\item \textbf{Heterogeneous Relation Awareness}. Our objective is to enhance the language model's proficiency in distinguishing between specific types of nodes within a heterogeneous context, taking into account the intricate relationships. This is achieved by leveraging the information encoded in the graph tokens. \\\vspace{-0.12in}

\item \textbf{Homogeneous Relation Awareness}. Our aim is to equip the model with the ability to establish a significant correspondence between sequences of graph tokens that belong to the same category and their corresponding natural language descriptions.

\end{itemize}




\noindent \textbf{Heterogeneous Graph Instruction}. In our graph instruction, we incorporate a heterogeneous subgraph generated through random neighbor sampling, accompanied by a question generated by a human. To enhance the diversity of the heterogeneous graph corpus, we conduct multiple samplings focusing on nodes from different types. Additionally, we introduce the \texttt{<graph>} token as a graph indicator within the human question. i) \emph{To achieve heterogeneous relation awareness}, we introduce the inter-type token matching task. This task involves providing the Language Model (LLM) with encoded sequences of graph tokens from different types, enabling it to differentiate between the various types. ii) \emph{For homogeneous relation awareness}, we design the intra-type matching task, where the LLM receives encoded sequences of graph tokens from a specific type, allowing it to establish correspondence with the relevant descriptions. Further details regarding the instruction template at this stage are illustrated in Table~\ref{tab:prompt} and Appendix Section~\ref{sec:inst_ds}.

\subsubsection{\bf Heterogeneity-aware Fine-Tuning}
To customize the reasoning abilities of the language model for specific downstream tasks on heterogeneous graphs, we propose Heterogeneity-aware Fine-Tuning. This approach entails conducting supervised learning with task-specific instructions following the initial instruction pre-training phase with heterogeneous graph corpus. It allows us to refine the LLM's performance and adapt it to the specific requirements of the targeted tasks on heterogeneous graphs.

In this stage, we incorporate a randomly sampled heterogeneous subgraph centered around the target node, along with a human-generated question. Given that the previous phase of instruction pre-training has already equipped the model with heterogeneous and homogeneous relation awareness, we design human questions that are rich in heterogeneity. These questions contain sequences of graph tokens from different types, indicated by multiple occurrences of the \texttt{<graph>} token. Additionally, the human question includes pertinent auxiliary information pertaining to the target node. The designs of these instructions are presented in Figure~\ref{tab:prompt}.


\vspace{-0.1in}
\subsection{Mixture-of-Thought (MoT) for Graph \\ Instruction Augmentation}
\begin{figure}
    \centering
\includegraphics[width=0.9\linewidth]{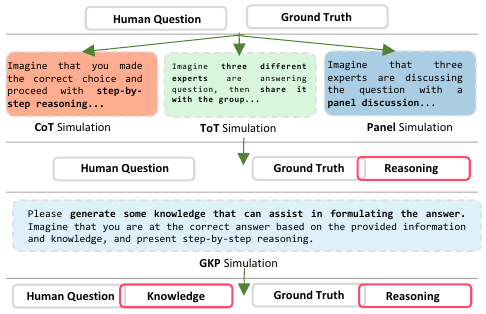}
    \vspace{-0.1in}
    \caption{Mixture-of-Thought (MoT) Augmentation}
    \label{fig:enter-label}
    \vspace{-0.25in}
\end{figure}
In practical scenarios of heterogeneous graph learning, data scarcity often poses a challenge. This is especially true when using heterogeneous graphs to model cold-start users/items in recommendation systems, where sparse user interaction data limits the availability of supervised signals. To address the issue of data sparsity, we propose enhancing our \model\ by incorporating augmented graph instructions. Drawing inspiration from previous works~\cite{COT_distill}, we introduce a novel method for instruction augmentation in the context of heterogeneous graph instruction tuning. This method utilizes prompt engineering techniques, particularly Mixture-of-Thought (MoT), to generate diverse and informative instructions. The goal is to effectively overcome the challenges posed by data scarcity. By incorporating augmented graph instructions, we expect our model enhancement to effectively handle data sparsity. \\\vspace{-0.12in}

\vspace{-0.05in}
\subsubsection{\bf Mixture-of-Thought (MoT) Prompting}
Our focus is on designing and optimizing prompts to effectively utilize language models \cite{cot, tot, panel_dis, GKP}. We employ several techniques to enhance language models: i) \textbf{Chain-of-Thought (CoT)} \cite{cot}: CoT prompts introduce intermediate steps, enabling complex reasoning and sophisticated capabilities. ii) \textbf{Tree-of-Thought (ToT)} \cite{tot}: ToT maintains a tree structure of coherent language sequences called thoughts. These thoughts serve as systematic intermediate steps for problem-solving. iii) \textbf{PanelGPT} \cite{panel_dis}: PanelGPT incorporates panel discussions among language models, enhancing the prompt engineering process through collaboration. iv) \textbf{Generated Knowledge Prompting (GKP)} \cite{GKP}: GKP involves incorporating additional knowledge into prompts for enhancement. By leveraging these techniques, our objective is to augment heterogeneous graph instructions, especially in scenarios with limited data availability.

\vspace{-0.05in}
\subsubsection{\bf Instruction Augmentation with Priori Knowledge}
We utilize seven instruction augmentation strategies, each generating seven augmented instructions for every question-answer pair, incorporating the characteristics of Mixture-of-Thought (MoT). However, closed-source language models such as ChatGPT may produce incorrect answers, resulting in flawed guidance. To overcome this issue, we propose incorporating prior knowledge, specifically the correct answer, into the prompt. It enables LLM to simulate generating the correct answer and produce intermediate reasoning steps using different MoT methods, as shown in Figure~\ref{tab:prompt} and Appendix.


\section{Evaluation}
\label{sec:eval}
To assess the effectiveness of our \model\ model, our experiments are designed to address the following research questions:

\begin{itemize}[leftmargin=*]

\item \textbf{RQ1:} How does the performance of our \model\ compare to that of baseline methods in both few-shot and zero-shot scenarios?

\item \textbf{RQ2:} To what extent do the key components of \model\ contribute to its overall performance across various settings?

\item \textbf{RQ3:} Can the \model's universal heterogeneity knowledge be leveraged to achieve graph in-context learning solely through graph instruction examples, without any model optimization?



\end{itemize}

\subsection{Experimental Settings}

\subsubsection{\bf Experimental Datasets}
The experiments were conducted on three benchmark datasets, \ie, IMDB~\cite{magnn}, DBLP~\cite{magnn}, and ACM~\cite{han}. \textbf{IMDB} is an extensive dataset that focuses on online movies and television programs. It encompasses 4278 movies, 2081 directors, and 5257 actors. Each movie is categorized into one of three classes: Action, Comedy, or Drama. \textbf{DBLP}, on the other hand, consists of 4057 authors, 14328 papers, 7723 terms, and 20 publication venues. This dataset was gathered from a computer science bibliography website and the authors are distributed among four research areas: Database, Data Mining, Artificial Intelligence, and Information Retrieval. Lastly, the \textbf{ACM} dataset comprises 3025 papers, 5835 authors, and 56 subjects. The papers are classified into three classes: Database, Wireless Communication, and Data Mining.

\vspace{-0.05in}
\subsubsection{\bf Evaluation Protocols}
To ensure consistency in the feature dimension of nodes across all datasets, we utilize a pre-trained Sentence-BERT to encode nodes of all types from each dataset into a standardized dimension. For the supervised few-shot node classification, we randomly select 1, 3, 5, 10, 20, 40, or 60 labeled nodes per class as our training set. Additionally, we reserve 1,000 nodes for validation and another 1,000 nodes for testing purposes. Our evaluation metrics encompass Micro-F1, Macro-F1, and AUC.

\vspace{-0.05in}
\subsubsection{\bf Compared Baseline Methods}
For our comprehensive performance comparison, we evaluate various state-of-the-art methods from three different categories: i) The first category consists of representative homogeneous graph neural networks, including \textbf{SAGE}~\cite{graphsage} and \textbf{GAT}~\cite{gat}. ii) The second category includes approaches utilizing message-passing mechanisms in heterogeneous graph neural networks. This category features models such as \textbf{HAN}~\cite{han}, \textbf{HGT}~\cite{HGT}, and \textbf{HetGNN}~\cite{hetgnn}. iii) The third category focuses on self-supervised techniques for heterogeneous graph learning. This category incorporates generative strategies like \textbf{HGMAE}~\cite{tian2023heterogeneous}, as well as contrastive schemes such as \textbf{DMGI}~\cite{park2020unsupervised} and \textbf{HeCo}~\cite{wang2021self}.

\vspace{-0.05in}
\subsubsection{\bf Implementation Details}
In Appendix Sec~\ref{sec:impl}, we offer comprehensive descriptions of the implementation details, including the datasets used, training hyperparameters, configurations of the base LLM, and more. These details provide a deeper understanding of our implementation approach.

\begin{table*}
\vspace{-0.15in}
\centering
\caption{Performance comparison on node classification tasks in both few-shot and zero-shot settings. However, since SSL methods focus on learning embeddings from downstream graphs, we excluded the zero-shot settings for them ("-").}\label{tab:performance_1}
\vspace{-0.15in}
\resizebox{1\textwidth}{!}{\begin{tabular}{c|c|c|c|clcccccc|cc} 
\hline
Datasets                     & Metric                 & train-on & test-on   & SAGE           & \multicolumn{1}{c}{GAT} & HAN           & HGT           & HetGNN           & DMGI          & HGMAE         & HeCo          & \textbf{\model-std}      & \textbf{\model-cot}       \\ 
\hline
\multirow{12}{*}{Supervised} & \multirow{4}{*}{Mi-F1} & IMDB-1   & IMDB-1000 & 0.4663$\pm$0.0025  & 0.4567$\pm$0.0122           & 0.4890$\pm$0.0271 & 0.4977$\pm$0.0186 & 0.4790$\pm$0.0134  & 0.4570$\pm$0.0126 & 0.3609$\pm$0.0145 & 0.3874$\pm$0.0159 & 0.5090$\pm$0.0073          & \textbf{0.5360$\pm$0.0065}  \\
                             &                        & IMDB-5   & IMDB-1000 & 0.5010$\pm$0.0051  & 0.5170$\pm$0.0029           & 0.4840$\pm$0.0094 & 0.5003$\pm$0.0093 & 0.5020$\pm$0.0045  & 0.4413$\pm$0.0173 & 0.3652$\pm$0.0062 & 0.3385$\pm$0.0169 & 0.6180$\pm$0.0027          & \textbf{0.6320$\pm$0.0085}  \\
                             &                        & IMDB-20  & IMDB-1000 & 0.5930$\pm$0.0093  & 0.6117$\pm$0.0012           & 0.5763$\pm$0.0046 & 0.5750$\pm$0.0065 & 0.5957$\pm$0.0054  & 0.5497$\pm$0.0256 & 0.4107$\pm$0.0106 & 0.3781$\pm$0.0148 & 0.6090$\pm$0.0255          & \textbf{0.6440$\pm$0.0075}  \\
                             &                        & IMDB-40  & IMDB-1000 & 0.6170$\pm$0.0112  & 0.6261$\pm$0.0015           & 0.6198$\pm$0.0025 & 0.5923$\pm$0.0040 & 0.6177$\pm$0.0046  & 0.5813$\pm$0.0033 & 0.3946$\pm$0.0067 & 0.3927$\pm$0.0134 & 0.6260$\pm$0.0057          & \textbf{0.6280$\pm$0.0071}  \\
\cline{2-14}
                             & \multirow{4}{*}{Ma-F1} & IMDB-1   & IMDB-1000 & 0.4425$\pm$0.0068  & 0.3974$\pm$0.0183           & 0.4229$\pm$0.0104 & 0.4020$\pm$0.0112 & 0.4456$\pm$0.0036  & 0.4083$\pm$0.0288 & 0.3573$\pm$0.0117 & 0.4023$\pm$0.0137 & 0.4986$\pm$0.0141          & \textbf{0.5247$\pm$0.0061}  \\
                             &                        & IMDB-5   & IMDB-1000 & 0.4613$\pm$0.0086  & 0.4767$\pm$0.0098           & 0.4695$\pm$0.0037 & 0.4676$\pm$0.0153 & 0.4677$\pm$0.0145  & 0.4254$\pm$0.0124 & 0.3500$\pm$0.0080 & 0.3468$\pm$0.0213 & 0.6111$\pm$0.0091          & \textbf{0.6243$\pm$0.0060}  \\
                             &                        & IMDB-20  & IMDB-1000 & 0.5953$\pm$0.0095  & 0.6121$\pm$0.0024           & 0.5756$\pm$0.0051 & 0.5723$\pm$0.0056 & 0.5969$\pm$0.0055  & 0.5495$\pm$0.0270 & 0.4065$\pm$0.0089 & 0.3904$\pm$0.0172 & 0.6068$\pm$0.0146          & \textbf{0.6398$\pm$0.0083}  \\
                             &                        & IMDB-40  & IMDB-1000 & 0.6182$\pm$0.0107  & 0.6254$\pm$0.0009           & 0.6224$\pm$0.0057 & 0.5909$\pm$0.0068 & 0.6234$\pm$0.0038  & 0.5786$\pm$0.0064 & 0.3866$\pm$0.0072 & 0.3988$\pm$0.0147 & \textbf{0.6265$\pm$0.0090} & 0.6237$\pm$0.0059           \\
\cline{2-14}
                             & \multirow{4}{*}{AUC}   & IMDB-1   & IMDB-1000 & 0.6079$\pm$0.0061  & 0.6151$\pm$0.0065           & 0.6234$\pm$0.0252 & 0.6249$\pm$0.0170 & 0.6107$\pm$0.0075  & 0.5780$\pm$0.0130 & 0.5274$\pm$0.0058 & 0.5712$\pm$0.0099 & 0.6565$\pm$0.0146          & \textbf{0.6685$\pm$0.0037}  \\
                             &                        & IMDB-5   & IMDB-1000 & 0.6309$\pm$0.0049  & 0.6372$\pm$0.0012           & 0.6102$\pm$0.0059 & 0.6197$\pm$0.0152 & 0.6290$\pm$0.0022  & 0.5832$\pm$0.0132 & 0.5262$\pm$0.0041 & 0.5067$\pm$0.0228 & 0.7308$\pm$0.0125          & \textbf{0.7310$\pm$0.0086}  \\
                             &                        & IMDB-20  & IMDB-1000 & 0.6976$\pm$0.0064  & 0.7122$\pm$0.0020           & 0.6815$\pm$0.0052 & 0.6801$\pm$0.0048 & 0.7005$\pm$0.0030  & 0.6657$\pm$0.0179 & 0.5766$\pm$0.0064 & 0.5541$\pm$0.0145 & 0.7227$\pm$0.0034          & \textbf{0.7424$\pm$0.0113}  \\
                             &                        & IMDB-40  & IMDB-1000 & 0.7171$\pm$0.0069  & 0.7210$\pm$0.0014           & 0.7204$\pm$0.0015 & 0.6970$\pm$0.0060 & 0.7145$\pm$0.0035  & 0.6860$\pm$0.0027 & 0.5488$\pm$0.0049 & 0.5653$\pm$0.0105 & 0.7323$\pm$0.0036          & \textbf{0.7331$\pm$0.0074}  \\
\hline
\multirow{24}{*}{Zero-shot}  & \multirow{4}{*}{Mi-F1} & IMDB-1   & DBLP-1000 & 0.2353$\pm$0.0372  & 0.1893$\pm$0.0373           & 0.2653$\pm$0.0203 & 0.2573$\pm$0.0519 & 0.2900$\pm$0.0638  & ~-~           & ~-~           & ~-~           & 0.3180$\pm$0.0072          & \textbf{0.3500$\pm$0.0073}  \\
                             &                        & IMDB-5   & DBLP-1000 & 0.2607$\pm$0.0082  & 0.2737$\pm$0.0176           & 0.2577$\pm$0.0094 & 0.2453$\pm$0.0458 & 0.2427$\pm$0.0452  & ~-~           & ~-~           & ~-~           & 0.3180$\pm$0.0044          & \textbf{0.3620$\pm$0.0047}  \\
                             &                        & IMDB-20  & DBLP-1000 & 0.2810$\pm$0.0289  & 0.2780$\pm$0.0033           & 0.2710$\pm$0.0000 & 0.2803$\pm$0.0208 & 0.2333$\pm$0.0353  & ~-~           & ~-~           & ~-~           & 0.3840$\pm$0.0088          & \textbf{0.4180$\pm$0.0083}  \\
                             &                        & IMDB-40  & DBLP-1000 & 0.2400$\pm$0.0324  & 0.2847$\pm$0.0053           & 0.2710$\pm$0.0000 & 0.2937$\pm$0.0005 & 0.2027$\pm$0.0345  & ~-~           & ~-~           & ~-~           & 0.3320$\pm$0.0087          & \textbf{0.3630$\pm$0.0045}  \\
\cline{2-14}
                             & \multirow{4}{*}{Ma-F1} & IMDB-1   & DBLP-1000 & 0.0963$\pm$0.0132  & 0.1169$\pm$0.0089           & 0.1047$\pm$0.0063 & 0.1016$\pm$0.0169 & 0.1778$\pm$0.0629  & ~-~           & ~-~           & ~-~           & 0.2048$\pm$0.0068          & \textbf{0.2472$\pm$0.0070}  \\
                             &                        & IMDB-5   & DBLP-1000 & 0.1042$\pm$0.0028  & 0.1291$\pm$0.0145           & 0.1024$\pm$0.0030 & 0.1138$\pm$0.0296 & 0.0971$\pm$0.0148  & ~-~           & ~-~           & ~-~           & 0.1917$\pm$0.0046          & \textbf{0.2773$\pm$0.0085}  \\
                             &                        & IMDB-20  & DBLP-1000 & 0.1448$\pm$0.0573  & 0.1274$\pm$0.0060           & 0.1066$\pm$0.0000 & 0.1143$\pm$0.0116 & 0.1008$\pm$0.0191  & ~-~           & ~-~           & ~-~           & 0.3142$\pm$0.0074          & \textbf{0.3733$\pm$0.0051}  \\
                             &                        & IMDB-40  & DBLP-1000 & 0.1068$\pm$0.0060  & 0.1588$\pm$0.0078           & 0.1066$\pm$0.0000 & 0.1268$\pm$0.0105 & 0.0984$\pm$0.0161  & ~-~           & ~-~           & ~-~           & 0.2331$\pm$0.0069          & \textbf{0.2912$\pm$0.0056}  \\
\cline{2-14}
                             & \multirow{4}{*}{AUC}   & IMDB-1   & DBLP-1000 & 0.4999$\pm$0.0001  & 0.4513$\pm$0.0295           & 0.5000$\pm$0.0000 & 0.5000$\pm$0.0000 & 0.5206$\pm$0.0306  & ~-~           & ~-~           & ~-~           & 0.5222$\pm$0.0069          & \textbf{0.5406$\pm$0.0040}  \\
                             &                        & IMDB-5   & DBLP-1000 & 0.4978$\pm$0.0030  & 0.4908$\pm$0.0078           & 0.5000$\pm$0.0000 & 0.5031$\pm$0.0043 & 0.4998$\pm$0.0003  & ~-~           & ~-~           & ~-~           & 0.5184$\pm$0.0081          & \textbf{0.5493$\pm$0.0091}  \\
                             &                        & IMDB-20  & DBLP-1000 & 0.5154$\pm$0.0213  & 0.4918$\pm$0.0020           & 0.5000$\pm$0.0000 & 0.5011$\pm$0.0016 & 0.4957$\pm$0.0060  & ~-~           & ~-~           & ~-~           & 0.5669$\pm$0.0041          & \textbf{0.5907$\pm$0.0089}  \\
                             &                        & IMDB-40  & DBLP-1000 & 0.5027$\pm$0.0031  & 0.4976$\pm$0.0021           & 0.5000$\pm$0.0000 & 0.5008$\pm$0.0006 & 0.4884$\pm$0.0164  & ~-~           & ~-~           & ~-~           & 0.5296$\pm$0.0070          & \textbf{0.5508$\pm$0.0086}  \\
\cline{2-14}
                             & \multirow{4}{*}{Mi-F1} & IMDB-1   & ACM-1000  & 0.3293$\pm$0.0418  & 0.3567$\pm$0.0053           & 0.3407$\pm$0.0111 & 0.3240$\pm$0.0014 & 0.3743$\pm$0.0434  & ~-~           & ~-~           & ~-~           & 0.4160$\pm$0.0106          & \textbf{0.4540$\pm$0.0089}  \\
                             &                        & IMDB-5   & ACM-1000  & 0.3820$\pm$0.0113  & 0.3787$\pm$0.0057           & 0.3630$\pm$0.0086 & 0.3160$\pm$0.0169 & 0.3583$\pm$0.0198  & ~-~           & ~-~           & ~-~           & 0.4580$\pm$0.0173          & \textbf{0.4880$\pm$0.0131}  \\
                             &                        & IMDB-20  & ACM-1000  & 0.2807$\pm$0.0074  & 0.3013$\pm$0.0188           & 0.3133$\pm$0.0031 & 0.3530$\pm$0.0000 & 0.2840$\pm$0.0226  & ~-~           & ~-~           & ~-~           & \textbf{0.5080$\pm$0.0129} & 0.5030$\pm$0.0064           \\
                             &                        & IMDB-40  & ACM-1000  & 0.3173$\pm$0.0005  & 0.2393$\pm$0.0144           & 0.2697$\pm$0.0194 & 0.3560$\pm$0.0099 & 0.3180$\pm$0.0016  & ~-~           & ~-~           & ~-~           & 0.4750$\pm$0.0149          & \textbf{0.5050$\pm$0.0077}  \\
\cline{2-14}
                             & \multirow{4}{*}{Ma-F1} & IMDB-1   & ACM-1000  & 0.2647$\pm$0.0269  & 0.2908$\pm$0.0131           & 0.2250$\pm$0.0416 & 0.1631$\pm$0.0005 & 0.3139$\pm$0.0468  & ~-~           & ~-~           & ~-~           & 0.3949$\pm$0.0078          & \textbf{0.4177$\pm$0.0124}  \\
                             &                        & IMDB-5   & ACM-1000  & 0.3208$\pm$0.0130  & 0.3009$\pm$0.0137           & 0.2782$\pm$0.0026 & 0.1969$\pm$0.0301 & 0.3087$\pm$0.0225  & ~-~           & ~-~           & ~-~           & 0.4336$\pm$0.0085          & \textbf{0.4510$\pm$0.0114}  \\
                             &                        & IMDB-20  & ACM-1000  & 0.2694$\pm$0.0091  & 0.2422$\pm$0.0098           & 0.2412$\pm$0.0050 & 0.2094$\pm$0.0501 & 0.2715$\pm$0.0181  & ~-~           & ~-~           & ~-~           & \textbf{0.4964$\pm$0.0075} & 0.4877$\pm$0.0070           \\
                             &                        & IMDB-40  & ACM-1000  & 0.3117$\pm$0.0017  & 0.2141$\pm$0.0071           & 0.2313$\pm$0.0132 & 0.2749$\pm$0.0122 & 0.3144$\pm$0.0017  & ~-~           & ~-~           & ~-~           & 0.4176$\pm$0.0116          & \textbf{0.4585$\pm$0.0089}  \\
\cline{2-14}
                             & \multirow{4}{*}{AUC}   & IMDB-1   & ACM-1000  & 0.4934$\pm$0.0247  & 0.5248$\pm$0.0038           & 0.5128$\pm$0.0086 & 0.5000$\pm$0.0000 & 0.5318$\pm$0.0295  & ~-~           & ~-~           & ~-~           & 0.5672$\pm$0.0040          & \textbf{0.5969$\pm$0.0082}  \\
                             &                        & IMDB-5   & ACM-1000  & 0.5433$\pm$0.0082  & 0.5415$\pm$0.0047           & 0.5282$\pm$0.0073 & 0.4950$\pm$0.0134 & 0.5256$\pm$0.0145  & ~-~           & ~-~           & ~-~           & 0.5991$\pm$0.0103          & \textbf{0.6224$\pm$0.0054}  \\
                             &                        & IMDB-20  & ACM-1000  & 0.4601$\pm$0.0048  & 0.4772$\pm$0.0137           & 0.4877$\pm$0.0029 & 0.5038$\pm$0.0053 & 0.4625$\pm$0.0163q & ~-~           & ~-~           & ~-~           & \textbf{0.6352$\pm$0.0094} & 0.6318$\pm$0.0068           \\
                             &                        & IMDB-40  & ACM-1000  & 0.4867$\pm$0.0013  & 0.4320$\pm$0.0108           & 0.4545$\pm$0.0146 & 0.5148$\pm$0.0043 & 0.4872$\pm$0.0006  & ~-~           & ~-~           & ~-~           & 0.6138$\pm$0.0047          & \textbf{0.6360$\pm$0.0051}  \\
\hline
\end{tabular}}
\vspace{-0.1in}
\end{table*}

\subsection{Overall Performance Comparison (RQ1)}
We performed node classification tasks on three datasets, exploring both few-shot and zero-shot settings. In the few-shot settings, our model was trained on the IMDB dataset with shot numbers ranging from 1 to 60, and evaluated on the IMDB test set of 1,000 samples~\cite{wang2021self, tian2023heterogeneous}. For the zero-shot settings, the model was trained on the IMDB dataset with the same shot numbers, and tested on separate test sets from the DBLP and ACM datasets, each containing 1,000 samples. To enable cross-dataset transferability in supervised heterogeneous Graph Neural Networks (GNNs), we unified node and edge categories, and utilized a classifier trained with transfer data to accommodate variations in class quantities across datasets.

For self-supervised methods focused on learning embeddings for downstream heterogeneous graph nodes, we excluded the zero-shot settings. The overall performance is partially shown in Table~\ref{tab:performance_1}, with detailed results in Table~\ref{tab:performance_app} in the appendix. "-std" and "-cot" notations represent the standard test prompt with direct answers and the prompt with a Chain-of-Thought (CoT) feature, respectively. These details provide insights into our node classification experiments in both supervised and zero-shot settings. \\\vspace{-0.12in}

\noindent \textbf{Superiority of \model\ in Few-Shot Settings}.
\model\ outperforms state-of-the-art baselines consistently in supervised settings, even with only one sample per class. The success can be attributed to our effective instruction-tuning on a large-scale heterogeneous graph corpus. This enables the LLM to extract valuable and transferable heterogeneous structural information from graph tokens, resulting in a significant performance boost in downstream tasks. Additionally, our proposed MoT graph instruction augmentation method enhances the LLM with diverse mixed reasoning capabilities without the need for additional supervision signals. As a result, it effectively mitigates the data scarcity in few-shot scenarios. \\\vspace{-0.12in}

\noindent \textbf{Zero-shot Superiority of \model}. 
In zero-shot settings, our \model\ surpasses baselines with significant improvements. Unlike traditional models constrained by training graph types, our approach leverages an in-context heterogeneous graph tokenizer. This tokenizer adapts tokenization based on the input graph, allowing the LLM to seamlessly combine graph tokens that capture higher-order structural features with its semantic understanding. As a result, our model effectively overcomes the limitation of graph heterogeneity shift, performing exceptionally well even in cross-domain scenarios. This showcases the remarkable adaptability of our \model. \\\vspace{-0.12in}

\noindent \textbf{Effectiveness of Mixture-of-Thought Augmentation}.
Through the implementation of the MoT approach, our model harnesses the varied reasoning capabilities of the formidable LLM (specifically, GPT-3.5) and seamlessly integrates them into our more compact language model. This integration serves to bolster our model's ability to effectively navigate data scarcity and elevate its performance in situations characterized by limited supervised signals. The MoT technique assumes a pivotal role in generating dynamic and diverse instructions, thereby offsetting the dearth of data and empowering our model to make notably precise predictions across both supervised and zero-shot settings.

\begin{table}
    \centering
    \caption{Ablation study of our \model.}\label{tab:ablation}
    \vspace{-0.15in}
    \resizebox{0.46\textwidth}{!}{\begin{tabular}{c|c|c|c|ccc|c} 
    \hline
    Datasets                     & Metric                 & train-on & test-on   & w/o S1 & w/o HG & w/o IA & \model             \\ 
    \cline{2-8}
    \multirow{12}{*}{Supervised} & \multirow{4}{*}{Mi-F1} & IMDB-1   & IMDB-1000 & fail        & 0.3740 &  0.4260   & \textbf{0.5360}  \\
                                 &                        & IMDB-3   & IMDB-1000 & fail        & 0.5000 &  0.4540   & \textbf{0.5730}  \\
                                 &                        & IMDB-10  & IMDB-1000 & fail        & 0.5660 &  0.4380   & \textbf{0.5810}  \\
                                 &                        & IMDB-20  & IMDB-1000 & fail        & 0.5640 &  0.5620   & \textbf{0.6440}  \\ 
    \cline{2-8}
                                 & \multirow{4}{*}{Ma-F1} & IMDB-1   & IMDB-1000 & fail        & 0.2433 &  0.3978  & \textbf{0.5247}  \\
                                 &                        & IMDB-3   & IMDB-1000 & fail        & 0.4969 &  0.4289  & \textbf{0.5591}  \\
                                 &                        & IMDB-10  & IMDB-1000 & fail        & 0.5619 &  0.3966  & \textbf{0.5762}  \\
                                 &                        & IMDB-20  & IMDB-1000 & fail        & 0.5636 &  0.5364  & \textbf{0.6398}  \\ 
    \cline{2-8}
                                 & \multirow{4}{*}{AUC}   & IMDB-1   & IMDB-1000 & fail        & 0.5195 &  0.6023  & \textbf{0.6685}  \\
                                 &                        & IMDB-3   & IMDB-1000 & fail        & 0.6340 &  0.6186  & \textbf{0.6935}  \\
                                 &                        & IMDB-10  & IMDB-1000 & fail        & 0.6790 &  0.5903  & \textbf{0.6875}  \\
                                 &                        & IMDB-20  & IMDB-1000 & fail        & 0.6891 &  0.6840  & \textbf{0.7424}  \\ 
    \hline
    \multirow{12}{*}{Zero-shot}  & \multirow{4}{*}{Mi-F1} & IMDB-1   & DBLP-1000 & fail        & 0.2980 & 0.2800  & \textbf{0.3500}  \\
                                 &                        & IMDB-3   & DBLP-1000 & fail        & 0.3430 & 0.3180  & \textbf{0.3660}  \\
                                 &                        & IMDB-10  & DBLP-1000 & fail        & 0.3640 & 0.3140  & \textbf{0.4020}  \\
                                 &                        & IMDB-20  & DBLP-1000 & fail        & 0.3920 & 0.2800  & \textbf{0.4180}  \\ 
    \cline{2-8}
                                 & \multirow{4}{*}{Ma-F1} & IMDB-1   & DBLP-1000 & fail        & 0.2444 & 0.2145  & \textbf{0.2472}  \\
                                 &                        & IMDB-3   & DBLP-1000 & fail        & 0.2768 & 0.2503  & \textbf{0.2814}  \\
                                 &                        & IMDB-10  & DBLP-1000 & fail        & 0.3211 & 0.2581  & \textbf{0.3386}  \\
                                 &                        & IMDB-20  & DBLP-1000 & fail        & 0.3689 & 0.1836  & \textbf{0.3733}  \\ 
    \cline{2-8}
                                 & \multirow{4}{*}{AUC}   & IMDB-1   & DBLP-1000 & fail        & 0.5275 & 0.5035   & \textbf{0.5406}  \\
                                 &                        & IMDB-3   & DBLP-1000 & fail        & 0.5422 & 0.5286   & \textbf{0.5524}  \\
                                 &                        & IMDB-10  & DBLP-1000 & fail        & 0.5636 & 0.5269   & \textbf{0.5777}  \\
                                 &                        & IMDB-20  & DBLP-1000 & fail        & 0.5834 & 0.4995   & \textbf{0.5907}  \\
    \hline
    \end{tabular}}
    \vspace{-0.15in}
    \end{table}

\subsection{Model Ablation Test (RQ2)}
To evaluate the proposed modules' effectiveness, we individually remove the key techniques in \model. The results are summarized in Table~\ref{tab:ablation}. Here are the ablated variants and the key conclusions:

\begin{itemize}[leftmargin=*]

  \item \textbf{Effect of Heterogeneous Graph Instruction-Tuning}. To validate the effectiveness of instruction tuning in the tuning stage on the large heterogeneous graph corpus, we generated the "w/o S1" variant by directly tuning the instructions solely on the downstream task data. Through experiments with different epoch settings (15, 50, and 100), we observed that models tuned solely on the downstream task data failed to provide complete and accurate answers in all cases. However, our \model\ achieved state-of-the-art performance within just 15 epochs across all settings. This success can be attributed to the fact that our \model\ learns from a vast heterogeneous graph context corpus, enabling it to understand and extract crucial structural information. As a result, in the second stage, our \model\ requires only a minimal amount of supervised data (even in a 1-shot scenario) to quickly align with the downstream task. Conversely, directly aligning LLMs with sparse supervised data proves to be challenging. \\\vspace{-0.12in}
  
  \item \textbf{Effect of In-Context Heterogeneous Graph Tokenizer}. We tested the necessity of incorporating heterogeneous graph structural information from our in-context tokenizer. By excluding the introduction of heterogeneous graph tokens and solely training the LLM's embeddings weights on the downstream data, we obtained a variant called "w/o HG". Our \model\ consistently outperformed this variant across different shot settings, especially in scenarios with limited samples (e.g., 1 or 3 shots). This improvement is attributed to the introduction of graph tokens, which enable the LLM to extract high-dimensional heterogeneous structural information from the in-context graph tokenizer. This enhanced understanding significantly improves the LLM's accuracy, particularly with sparse supervised signals.

  \item \textbf{Effect of MoT Instruction Augmentation}. To verify the effectiveness of the MoT graph instruction augmentation strategy, we trained the variant "-IA" using only direct-answer instructions. Results showed a significant drop in model performance without instruction augmentation, highlighting its importance in tackling the scarcity of labels in downstream tasks. Additionally, \model's superior performance in zero-shot settings can be attributed to its enhanced reasoning ability, acquired through training with diverse reasoning instructions. This improved capacity enables effective cross-dataset and cross-domain transfer.
\end{itemize}

\subsection{Graph In-Context Learning (RQ3)}
\begin{figure}[t]
  \vspace{-0.15in}
  \centering
    
  \subfigure[\bf IMDB-IMDB@Mi-F1, Ma-F1]{
      \centering
      \includegraphics[width=0.22\textwidth]{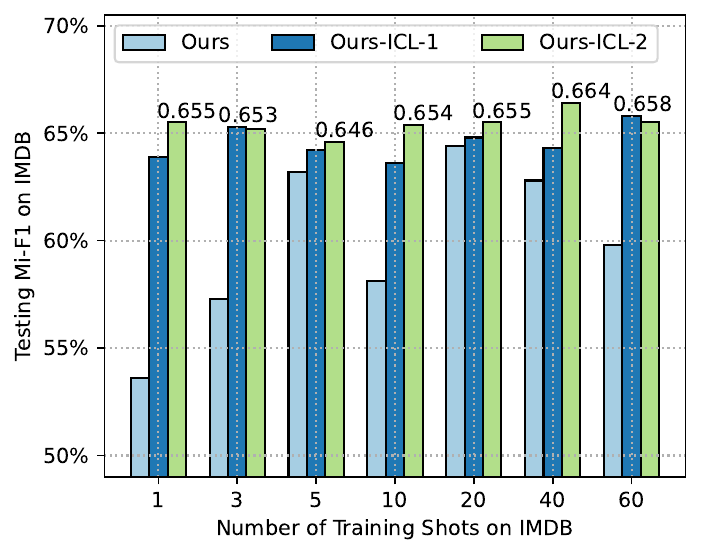}
      \includegraphics[width=0.22\textwidth]{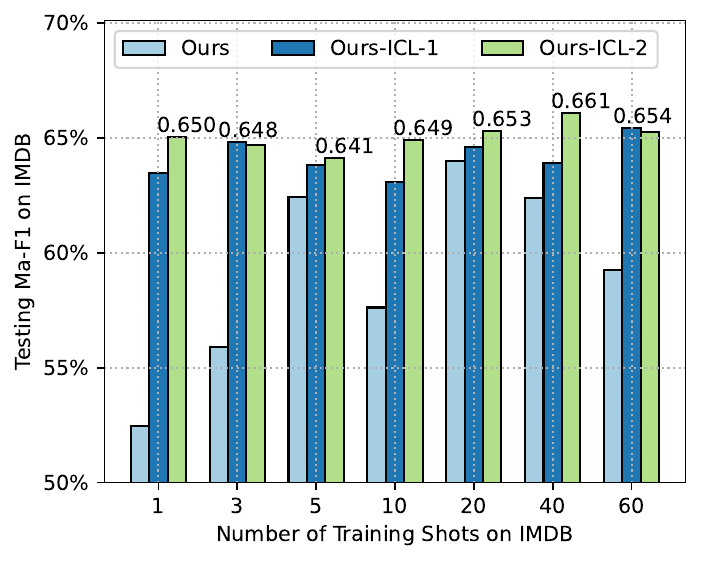}
  }\vspace{-0.15in}
  \subfigure[\bf IMDB-DBLP@Mi-F1, Ma-F1]{
      \centering
      \includegraphics[width=0.22\textwidth]{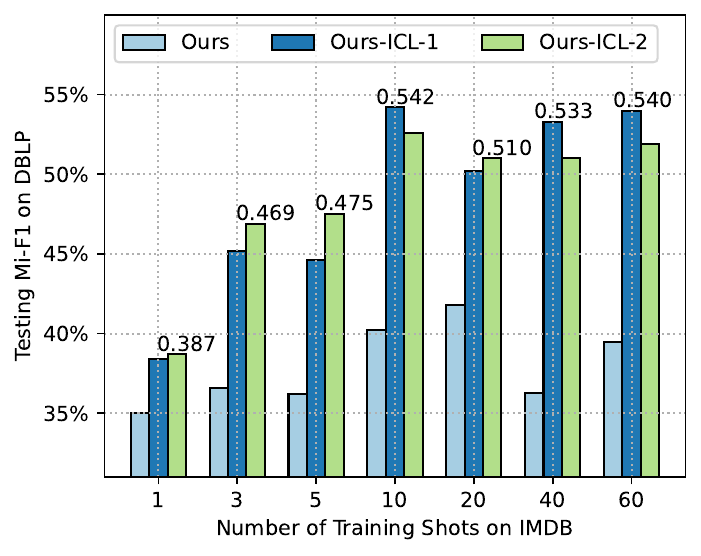}
      \includegraphics[width=0.22\textwidth]{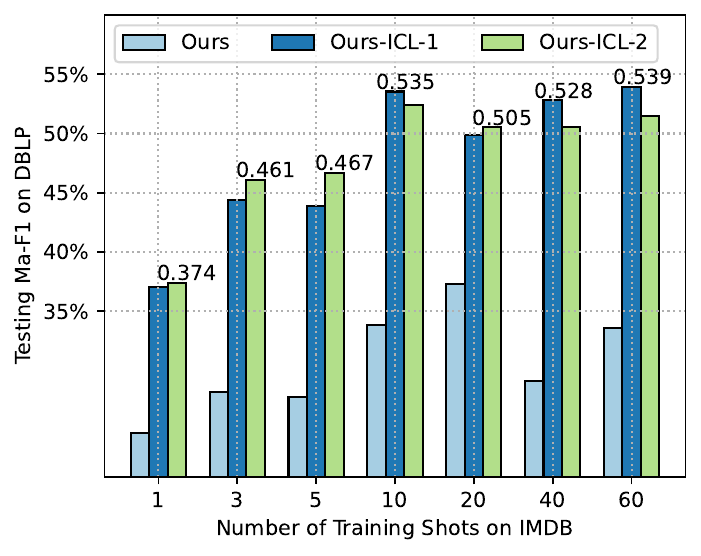}
  }\vspace{-0.15in}
  \subfigure[\bf IMDB-ACM@Mi-F1, Ma-F1]{
      \centering
      \includegraphics[width=0.22\textwidth]{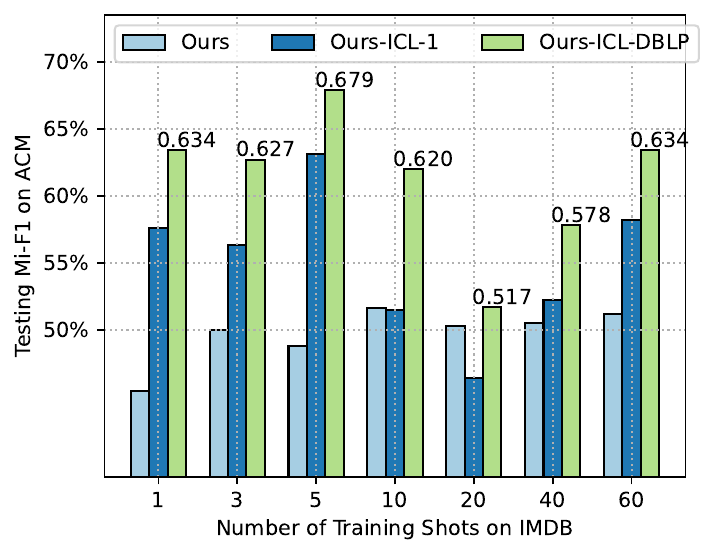}
      \includegraphics[width=0.22\textwidth]{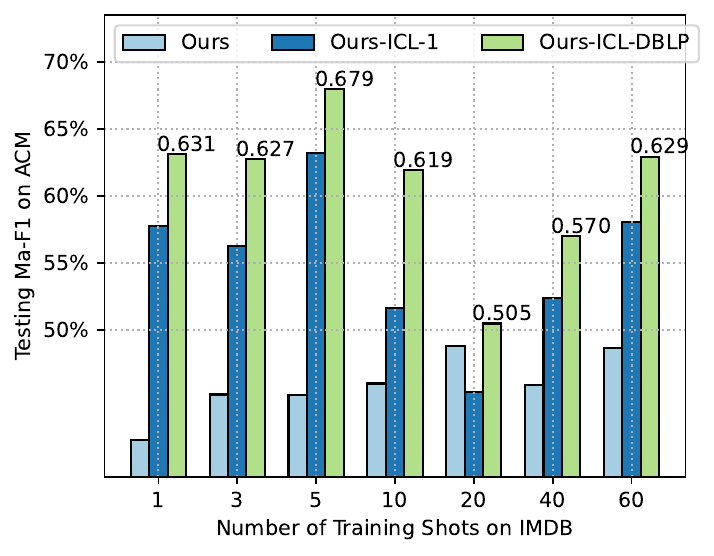}
  }\vspace{-0.2in}
      \caption{Graph In-Context Learning of our \model. }\label{fig:graph_icl}
  \vspace{-0.25in}
\end{figure}

In-context learning (ICL)\cite{in_context} is a method for adapting large language models (LLMs) to new tasks without gradient updates, using a prompt with task examples. In this subsection, we explore the impact of Graph In-Context Learning on improving \model's performance. We conduct comprehensive tests by adding prefatory examples from the training set to models trained with different shots of IMDB data. We randomly sampled training examples corresponding to the test data. "-ICL-1" and "-ICL-2" denote one and two prefatory examples, respectively. "-ICL-DBLP" signifies the inclusion of DBLP examples before the ACM test prompt. The results, depicted in Figure\ref{fig:graph_icl}, reveal the following observations: \\\vspace{-0.12in}

\noindent\textbf{1-shot Beat 60-shot with Graph ICL in \model.} 
Results show that, even with just a single example, most 1-shot models using Graph ICL consistently outperform 60-shot models without further training in both supervised and zero-shot settings. Increasing the number of examples enhances the effect of in-context learning. This improvement can be attributed to \model's two-stage instruction tuning process, which enables it to understand and analyze heterogeneous graph tokens, benefiting downstream tasks. By providing question-and-answer examples with graph tokens, the model gains a deeper understanding of the graph-text relationship. Analyzing and emulating these examples leads to more accurate responses. \\\vspace{-0.12in}

\noindent \textbf{Enhanced Transferability with our Graph ICL.} 
The advantages of our Graph ICL in \model\ are particularly evident in zero-shot transfer scenarios. This indicates that the Graph ICL approach significantly improves \model's transferability without the need to optimize model parameters. Our \model\ does not simply overfit to a single dataset but develops the ability to analyze text alongside heterogeneous graph tokens. By incorporating graph examples from other datasets, the model effortlessly transfers this analytical capability, demonstrating strong transfer learning capacity. \\\vspace{-0.12in}

\noindent \textbf{Benefit of Irrelevant Graph Examples.} 
We experimented with \model\ with irrelevant graph examples, like using DBLP Q\&A examples for testing on the ACM dataset. Surprisingly, using DBLP graph examples yielded the best results. Despite different target tasks, our \model\ effectively leverages in-context information from heterogeneous graph tokens, enhancing downstream tasks. This confirms that our \model\ learns valuable information from heterogeneous graph structures, rather than relying solely on text. Using ACM's own examples did not perform as well due to a deficiency in encoding the ACM graph in the alignment and stage 1 process. However, the DBLP examples mitigated this issue to some extent.

\vspace{-0.05in}
\subsection{Case Study}
We perform a case study to showcase our \model's robust generalization in understanding complex graph structures with diverse nodes and connections. Our model generates graph-aware predictions and responses, demonstrating its profound comprehension and awareness of graph-related aspects. Furthermore, we validate the positive impact of our MoT instruction augmentation. For more comprehensive details, please refer to Appendix Section~\ref{sec:case_study}.
\section{Related Work}
\label{sec:relate}

\noindent \textbf{Heterogeneous Graph Neural Networks}.
Heterogeneous Graph Neural Networks (HGNNs) capture complex relationships and diverse semantics among entities in a heterogeneous graph \cite{fan2019metapath, chen2023heterogeneous, hetgnn}. They use specialized message functions and aggregation rules to model relation heterogeneity. Existing models, such as MAGNN \cite{fan2019metapath} and HetGNN \cite{hetgnn}, leverage metapaths to capture composite relations and guide neighbor selection. Heterogeneous graph convolution frameworks like HeteGCN \cite{ragesh2021hetegcn} draws inspiration from graph convolutional networks. Heterogeneous graph attention networks, including HAN \cite{wang2019heterogeneous}, HGT \cite{HGT}, and HGAT \cite{linmei2019heterogeneous}, use attention mechanisms to effectively capture and aggregate information from different node types. However, most HGNNs require sufficient labels to learn accurate graph representations. \\\vspace{-0.12in}


\noindent \textbf{Heterogeneous Graph Self-Supervised Learning}. 
Recent research has addressed the limited availability of labeled data by incorporating self-supervised learning techniques into heterogeneous graph modeling~\cite{wang2021selfwww, jing2022x}. Contrastive and generative approaches have proven effective in augmenting data. Contrastive learning methods like DMGI~\cite{park2020unsupervised} and HeCo~\cite{wang2021self} bring similar instances closer and push dissimilar instances apart in a latent space, capturing relevant patterns and structure. Generative learning approaches such as HGMAE~\cite{tian2023heterogeneous} use masked autoencoders to reconstruct heterogeneous graphs. However, these approaches still have limitations in handling relation heterogeneity shift across downstream tasks, potentially leading to poor generalization ability. \\\vspace{-0.12in}

\noindent \textbf{Large Language Models for Graph Data}. 
Recent research has combined large language models (LLMs) and graph models to understand complex relationships in graph data~\cite{wei2023llmrec,ye2023natural,chai2023graphllm,ren2023representation}. Two primary approaches integrate graph structural information: utilizing textual prompts and incorporating graph embeddings as input tokens. Chen et al.~\cite{potential_llm} craft tailored prompts for graph learning tasks, while InstructGLM~\cite{ye2023natural} and GraphGPT~\cite{tang2023graphgpt} propose to integrate prompt instructions with graph embeddings for fine-tuning LLM. Moreover, advancements have introduced LLMs to improve the reasoning capabilities of models when working with graph-structured data. Prominent examples include the works of Fatemi et al.~\cite{fatemi2023talk} and Chai et al.~\cite{chai2023graphllm}. However, existing LLM-enhanced graph models have primarily focused on homogeneous graphs, overlooking the inherent heterogeneity in real-world graphs. This calls for further exploration of heterogeneous graph language models with strong generalization abilities across diverse downstream tasks.
\section{Conclusion}
\label{sec:conclusion}

This work introduce \model, a general and versatile graph model that offers the ability to learn from diverse heterogeneous graphs without the need for downstream fine-tuning processes. To address distribution shifts in heterogeneity, we propose an in-context heterogeneous graph tokenizer that captures semantic relationships across different heterogeneous graphs, facilitating seamless model adaptation. By incorporating the heterogeneity-aware graph instructions into our \model, the model becomes proficient in comprehending intricate relation heterogeneity and accurately discerning between various types of graph tokens. Our proposed framework has undergone extensive evaluations across diverse scenarios, demonstrating outstanding generalization performance.

\clearpage

\bibliographystyle{abbrv}
\balance
\bibliography{refs}

\begin{thebibliography}{10}

\bibitem{chai2023graphllm}
Z.~Chai, T.~Zhang, L.~Wu, K.~Han, X.~Hu, X.~Huang, and Y.~Yang.
\newblock Graphllm: Boosting graph reasoning ability of large language model.
\newblock {\em arXiv preprint arXiv:2310.05845}, 2023.

\bibitem{chen2023heterogeneous}
M.~Chen, C.~Huang, L.~Xia, W.~Wei, Y.~Xu, and R.~Luo.
\newblock Heterogeneous graph contrastive learning for recommendation.
\newblock In {\em WSDM}, pages 544--552, 2023.

\bibitem{potential_llm}
Z.~Chen, H.~Mao, H.~Li, et~al.
\newblock Exploring the potential of large language models (llms) in learning on graphs.
\newblock {\em CoRR}, abs/2307.03393, 2023.

\bibitem{dong2017metapath2vec}
Y.~Dong, N.~V. Chawla, and A.~Swami.
\newblock metapath2vec: Scalable representation learning for heterogeneous networks.
\newblock In {\em KDD}, pages 135--144, 2017.

\bibitem{el2022twhin}
A.~El-Kishky, T.~Markovich, S.~Park, C.~Verma, B.~Kim, R.~Eskander, Y.~Malkov, F.~Portman, S.~Samaniego, Y.~Xiao, et~al.
\newblock Twhin: Embedding the twitter heterogeneous information network for personalized recommendation.
\newblock In {\em KDD}, pages 2842--2850, 2022.

\bibitem{fan2019metapath}
S.~Fan, J.~Zhu, X.~Han, C.~Shi, L.~Hu, B.~Ma, and Y.~Li.
\newblock Metapath-guided heterogeneous graph neural network for intent recommendation.
\newblock In {\em KDD}, pages 2478--2486, 2019.

\bibitem{fatemi2023talk}
B.~Fatemi, J.~Halcrow, and B.~Perozzi.
\newblock Talk like a graph: Encoding graphs for large language models.
\newblock {\em arXiv preprint arXiv:2310.04560}, 2023.

\bibitem{magnn}
X.~Fu, J.~Zhang, Z.~Meng, and I.~King.
\newblock {MAGNN:} metapath aggregated graph neural network for heterogeneous graph embedding.
\newblock In {\em {WWW}}, pages 2331--2341. {ACM} / {IW3C2}, 2020.

\bibitem{graphsage}
W.~L. Hamilton, Z.~Ying, and J.~Leskovec.
\newblock Inductive representation learning on large graphs.
\newblock In {\em {NeurIPS}}, pages 1024--1034, 2017.

\bibitem{HGT}
Z.~Hu, Y.~Dong, K.~Wang, and Y.~Sun.
\newblock Heterogeneous graph transformer.
\newblock In {\em {WWW}}, pages 2704--2710. {ACM} / {IW3C2}, 2020.

\bibitem{hwang2020self}
D.~Hwang, J.~Park, S.~Kwon, K.~Kim, J.-W. Ha, and H.~J. Kim.
\newblock Self-supervised auxiliary learning with meta-paths for heterogeneous graphs.
\newblock {\em NeurIPS}, 33:10294--10305, 2020.

\bibitem{jin2021heterogeneous}
D.~Jin, C.~Huo, C.~Liang, and L.~Yang.
\newblock Heterogeneous graph neural network via attribute completion.
\newblock In {\em WWW}, pages 391--400, 2021.

\bibitem{jing2022x}
B.~Jing, S.~Feng, Y.~Xiang, X.~Chen, Y.~Chen, and H.~Tong.
\newblock X-goal: multiplex heterogeneous graph prototypical contrastive learning.
\newblock In {\em CIKM}, pages 894--904, 2022.

\bibitem{linmei2019heterogeneous}
H.~Linmei, T.~Yang, C.~Shi, H.~Ji, and X.~Li.
\newblock Heterogeneous graph attention networks for semi-supervised short text classification.
\newblock In {\em EMNLP}, pages 4821--4830, 2019.

\bibitem{GKP}
J.~Liu, A.~Liu, X.~Lu, S.~Welleck, P.~West, R.~L. Bras, Y.~Choi, and H.~Hajishirzi.
\newblock Generated knowledge prompting for commonsense reasoning.
\newblock In {\em {ACL} {(1)}}, pages 3154--3169. Association for Computational Linguistics, 2022.

\bibitem{lv2021we}
Q.~Lv, M.~Ding, Q.~Liu, Y.~Chen, W.~Feng, S.~He, C.~Zhou, J.~Jiang, Y.~Dong, and J.~Tang.
\newblock Are we really making much progress? revisiting, benchmarking and refining heterogeneous graph neural networks.
\newblock In {\em KDD}, pages 1150--1160, 2021.

\bibitem{ma2023single}
A.~Ma, X.~Wang, J.~Li, C.~Wang, T.~Xiao, Y.~Liu, H.~Cheng, J.~Wang, Y.~Li, Y.~Chang, et~al.
\newblock Single-cell biological network inference using a heterogeneous graph transformer.
\newblock {\em Nature Communications}, 14(1):964, 2023.

\bibitem{in_context}
S.~Min, X.~Lyu, A.~Holtzman, M.~Artetxe, M.~Lewis, H.~Hajishirzi, and L.~Zettlemoyer.
\newblock Rethinking the role of demonstrations: What makes in-context learning work?
\newblock In {\em {EMNLP}}, pages 11048--11064, 2022.

\bibitem{park2020unsupervised}
C.~Park, D.~Kim, J.~Han, and H.~Yu.
\newblock Unsupervised attributed multiplex network embedding.
\newblock In {\em AAAI}, volume~34, pages 5371--5378, 2020.

\bibitem{CLIP}
A.~Radford, J.~W. Kim, C.~Hallacy, et~al.
\newblock Learning transferable visual models from natural language supervision.
\newblock In {\em International Conference on Machine Learning (ICML)}, pages 8748--8763. {PMLR}, 2021.

\bibitem{ragesh2021hetegcn}
R.~Ragesh, S.~Sellamanickam, A.~Iyer, R.~Bairi, and V.~Lingam.
\newblock Hetegcn: heterogeneous graph convolutional networks for text classification.
\newblock In {\em WSDM}, pages 860--868, 2021.

\bibitem{sent_bert}
N.~Reimers and I.~Gurevych.
\newblock Sentence-bert: Sentence embeddings using siamese bert-networks.
\newblock In {\em EMNLP}. Association for Computational Linguistics, 11 2019.

\bibitem{ren2023representation}
X.~Ren, W.~Wei, L.~Xia, L.~Su, S.~Cheng, J.~Wang, D.~Yin, and C.~Huang.
\newblock Representation learning with large language models for recommendation.
\newblock {\em arXiv preprint arXiv:2310.15950}, 2023.

\bibitem{COT_distill}
K.~Shridhar, A.~Stolfo, and M.~Sachan.
\newblock Distilling reasoning capabilities into smaller language models.
\newblock In {\em ACL}, pages 7059--7073, 2023.

\bibitem{panel_dis}
H.~Sun, A.~H{\"u}y{\"u}k, and M.~van~der Schaar.
\newblock Query-dependent prompt evaluation and optimization with offline inverse rl.
\newblock {\em arXiv e-prints}, pages arXiv--2309, 2023.

\bibitem{tang2023graphgpt}
J.~Tang, Y.~Yang, W.~Wei, L.~Shi, L.~Su, S.~Cheng, D.~Yin, and C.~Huang.
\newblock Graphgpt: Graph instruction tuning for large language models, 2023.

\bibitem{tian2023heterogeneous}
Y.~Tian, K.~Dong, C.~Zhang, C.~Zhang, and N.~V. Chawla.
\newblock Heterogeneous graph masked autoencoders.
\newblock In {\em AAAI}, volume~37, pages 9997--10005, 2023.

\bibitem{gat}
P.~Velickovic, G.~Cucurull, A.~Casanova, A.~Romero, et~al.
\newblock Graph attention networks.
\newblock In {\em {ICLR} (Poster)}. OpenReview.net, 2018.

\bibitem{wang2021relational}
H.~Wang, H.~Ren, and J.~Leskovec.
\newblock Relational message passing for knowledge graph completion.
\newblock In {\em KDD}, pages 1697--1707, 2021.

\bibitem{wang2021selfwww}
P.~Wang, K.~Agarwal, C.~Ham, S.~Choudhury, and C.~K. Reddy.
\newblock Self-supervised learning of contextual embeddings for link prediction in heterogeneous networks.
\newblock In {\em WWW}, pages 2946--2957, 2021.

\bibitem{wang2022survey}
X.~Wang, D.~Bo, C.~Shi, S.~Fan, Y.~Ye, and S.~Y. Philip.
\newblock A survey on heterogeneous graph embedding: methods, techniques, applications and sources.
\newblock {\em Transactions on Big Data (TBD)}, 9(2):415--436, 2022.

\bibitem{wang2019heterogeneous}
X.~Wang, H.~Ji, C.~Shi, B.~Wang, Y.~Ye, P.~Cui, and P.~S. Yu.
\newblock Heterogeneous graph attention network.
\newblock In {\em WWW}, pages 2022--2032, 2019.

\bibitem{han}
X.~Wang, H.~Ji, C.~Shi, B.~Wang, Y.~Ye, et~al.
\newblock Heterogeneous graph attention network.
\newblock In {\em {WWW}}, pages 2022--2032. {ACM}, 2019.

\bibitem{wang2021self}
X.~Wang, N.~Liu, H.~Han, and C.~Shi.
\newblock Self-supervised heterogeneous graph neural network with co-contrastive learning.
\newblock In {\em KDD}, pages 1726--1736, 2021.

\bibitem{cot}
J.~Wei, X.~Wang, D.~Schuurmans, M.~Bosma, B.~Ichter, F.~Xia, E.~H. Chi, Q.~V. Le, and D.~Zhou.
\newblock Chain-of-thought prompting elicits reasoning in large language models.
\newblock In {\em NeurIPS}, 2022.

\bibitem{wei2022contrastive}
W.~Wei, C.~Huang, L.~Xia, Y.~Xu, J.~Zhao, and D.~Yin.
\newblock Contrastive meta learning with behavior multiplicity for recommendation.
\newblock In {\em WSDM}, pages 1120--1128, 2022.

\bibitem{wei2023llmrec}
W.~Wei, X.~Ren, J.~Tang, Q.~Wang, L.~Su, S.~Cheng, J.~Wang, D.~Yin, and C.~Huang.
\newblock Llmrec: Large language models with graph augmentation for recommendation.
\newblock {\em arXiv preprint arXiv:2311.00423}, 2023.

\bibitem{G2P2}
Z.~Wen and Y.~Fang.
\newblock Augmenting low-resource text classification with graph-grounded pre-training and prompting.
\newblock In {\em SIGIR}, 2023.

\bibitem{xuan2023knowledge}
H.~Xuan, Y.~Liu, B.~Li, and H.~Yin.
\newblock Knowledge enhancement for contrastive multi-behavior recommendation.
\newblock In {\em WSDM}, pages 195--203, 2023.

\bibitem{yang2020heterogeneous}
C.~Yang, Y.~Xiao, Y.~Zhang, Y.~Sun, and J.~Han.
\newblock Heterogeneous network representation learning: A unified framework with survey and benchmark.
\newblock {\em Transactions on Knowledge and Data Engineering (TKDE)}, 34(10):4854--4873, 2020.

\bibitem{yang2022self}
Y.~Yang, Z.~Guan, Z.~Wang, W.~Zhao, C.~Xu, W.~Lu, and J.~Huang.
\newblock Self-supervised heterogeneous graph pre-training based on structural clustering.
\newblock {\em NeurIPS}, 35:16962--16974, 2022.

\bibitem{tot}
S.~Yao, D.~Yu, J.~Zhao, I.~Shafran, T.~L. Griffiths, Y.~Cao, and K.~Narasimhan.
\newblock Tree of thoughts: Deliberate problem solving with large language models.
\newblock {\em CoRR}, abs/2305.10601, 2023.

\bibitem{ye2023natural}
R.~Ye, C.~Zhang, R.~Wang, S.~Xu, and Y.~Zhang.
\newblock Natural language is all a graph needs.
\newblock {\em arXiv preprint arXiv:2308.07134}, 2023.

\bibitem{hetgnn}
C.~Zhang, D.~Song, C.~Huang, A.~Swami, and N.~V. Chawla.
\newblock Heterogeneous graph neural network.
\newblock In {\em {KDD}}, pages 793--803. {ACM}, 2019.

\bibitem{zhao2021heterogeneous}
J.~Zhao, X.~Wang, C.~Shi, B.~Hu, G.~Song, and Y.~Ye.
\newblock Heterogeneous graph structure learning for graph neural networks.
\newblock In {\em AAAI}, volume~35, pages 4697--4705, 2021.

\end{thebibliography}

\clearpage
\appendix
\section{Appendix}
\label{tab:appendix}
In the supplementary materials, we provide detailed information pertaining to our experiments. First, in Section~\ref{sec:data_ds}, we present the statistical information of the datasets used. Next, in Section~\ref{sec:baseline_ds}, we provide a comprehensive description of the baselines employed. In Section~\ref{sec:type_ds}, we outline the specific descriptions of nodes and edges in our text-enriched heterogeneity representations. Section~\ref{sec:inst_ds} elaborates on the templates for instructions and prompts used in our \model, including a detailed explanation of MoT graph instruction augmentation, an instruction construction template for two-stage instruction tuning, and an instruction construction template for graph in-context learning. Additionally, Section~\ref{sec:exp_app} presents additional experimental results, covering implementation details of our model, model case studies, overall performance, and comprehensive results of graph in-context learning.

\subsection{Detailed Statistics of Datasets}\label{sec:data_ds}
In Table~\ref{tab:data_ds}, we present the statistical information of the datasets used in our experiments, where the types of the target nodes in each heterogeneous graph are highlighted with an \underline{underline}.
\begin{table}[h]
\centering
\vspace{-0.1in}
\caption{Detailed statistics of utilzed datasets.}\label{tab:data_ds}
\vspace{-0.1in}
\resizebox{0.46\textwidth}{!}{\begin{tabular}{c|c|c|c|c} 
\hline
Dataset               & \# Nodes       & \# Edges    & Metapaths & \# Classes          \\ 
\hline
\multirow{4}{*}{ACM}  & \underline{Paper}: 3025    & P-A: 9949   & PAP       & \multirow{4}{*}{3}  \\
                      & Author: 5959   & P-S: 3025   & PSP       &                     \\
                      & Subject: 56    & P-T: 255619 & PTP       &                     \\
                      & Term: 1902     & P-P: 5343   &           &                     \\ 
\hline
\multirow{4}{*}{DBLP} & \underline{Author}: 4057   & P-A: 19645  & PAP       & \multirow{4}{*}{4}  \\
                      & Paper: 14328   & P-C: 14328  & APCPA     &                     \\
                      & Term: 7723     & P-T: 85810  & APTPA     &                     \\
                      & Conference: 20 &             &           &                     \\ 
\hline
\multirow{3}{*}{IMDB} & \underline{Movie}: 4278    & M-D: 4278   & MAM       & \multirow{3}{*}{3}  \\
                      & Director: 2081 & M-A: 12828  & MDM       &                     \\
                      & Actor: 5257    &             &           &                     \\
\hline
\end{tabular}}
\vspace{-0.1in}
\end{table}

\subsection{Detailed Descriptions of Baselines}\label{sec:baseline_ds}
To conduct a thorough evaluation, our baseline set includes the following methods, which are presented below:

\noindent\textbf{1) Homogeneous Graph Neural Networks}
\begin{itemize}[leftmargin=*]
\item \textbf{SAGE}~\cite{graphsage}:It was developed to facilitate the learning of inductive representations on large-scale homogeneous graphs, allowing for the generation of highly effective node embeddings for unseen data. However, we made adaptations and modifications to tailor it specifically to the unique demands of heterogeneous graphs. \\\vspace{-0.12in}
\item \textbf{GAT}~\cite{gat}: It utilizes graph attention networks, which overcome the limitations of graph convolutional networks (GCN) by incorporating masked self-attention layers. What sets this method apart is its ability to selectively weigh the aggregated information from different nodes, thereby enhancing the message passing mechanism and refining the overall process. \\\vspace{-0.12in}
\end{itemize}

\noindent\textbf{2) Heterogeneous Graph Neural Encoders.}
\begin{itemize}[leftmargin=*]
\item \textbf{HAN}~\cite{han}: It is a heterogeneous graph neural encoder that incorporates hierarchical attention mechanisms, including node-level and semantic-level attentions, to address the complexity of heterogeneous graphs with various types of nodes and links, thereby improving the representation and interpretability of node embeddings through feature aggregation based on meta-path neighbors. \\\vspace{-0.12in}
\item \textbf{HGT}~\cite{HGT}: 
HGT is an advanced graph neural network framework designed to model the complexities of large-scale heterogeneous graphs, featuring type-dependent parameters for nodes and edges to enable heterogeneous attention, a relative temporal encoding to capture dynamic relationships, and an efficient graph sampling algorithm for scalable training. \\\vspace{-0.12in}
\item \textbf{HetGNN}~\cite{hetgnn}: HetGNN is a powerful heterogeneous graph neural network model that seamlessly integrates both the structural information and diverse content attributes of nodes. It achieves this by employing a two-module architecture for feature aggregation and incorporating a well-designed random walk sampling process. This comprehensive approach enables HetGNN to generate meaningful and informative node embeddings. \\\vspace{-0.12in}
\end{itemize}
\noindent\textbf{3) SSL-enhanced Heterogeneous Graph Learning Approaches.}
\begin{itemize}[leftmargin=*]
\item \textbf{HGMAE}~\cite{tian2023heterogeneous}: HGMAE is a generative SSL approach that addresses the challenges of capturing complex structures, incorporating diverse node attributes, and encoding node positions in heterogeneous graphs. It achieves this through innovative masking techniques and tailored training strategies, utilizing a heterogeneous graph masked autoencoder. HGMAE efficiently learns to generate meaningful representations while effectively preserving the rich information present in heterogeneous graphs. \\\vspace{-0.12in}
\item \textbf{DMGI}~\cite{park2020unsupervised}: It is an effective unsupervised network embedding method for attributed multiplex networks that maximizes mutual information between graph patches and a global representation, integrating multiple relation-type embeddings with a consensus regularization framework and a universal discriminator, further enhanced by an attention mechanism to weigh relation types. \\\vspace{-0.12in}
\item \textbf{HeCo}~\cite{wang2021self}: 
This is a self-supervised heterogeneous graph neural network framework that employs a co-contrastive learning mechanism across two views (network schema and meta-path) to capture both local and high-order structures, with a view mask mechanism for effective cross-view supervision, enhanced by extensions for generating higher-quality negative samples. \\\vspace{-0.12in}
\end{itemize}
\subsection{Detailed Descriptions of Text-Enriched Heterogeneity Representations }\label{sec:type_ds}
In Table~\ref{tab:des_type} and~\ref{tab:des_type_2}, we present diverse descriptions for different node and edge types across three datasets, \ie, IMDB, DBLP and ACM.
\begin{table*}
  \centering
  \caption{Detailed Descriptions of Text-Enriched Heterogeneity Representations for IMDB and DBLP.}
  \vspace{-0.15in}
  \label{tab:des_type}
  \resizebox{1\linewidth}{!}{
  \begin{tabular}{p{3.3cm}p{15cm}}
    \toprule
    Node (Edge) Type & Sets of Descriptions  \\
    \midrule
    \multicolumn{2}{c}{(a) IMDB} \\
    \cmidrule(rl){1-2}
    "Movie"
    & 
    \{"This node represents a movie","This is an action movie","This is a comedy movie","This is a drama movie"\} \\
    "Director"
    & 
    \{"This node represents a director",
"This is an action film director",
"The director specializes in action",
"This is a comedy film director",
"The director specializes in comedy",
"This is a drama film director",
"The director specializes in dram"\} \\
"Actor"
&
\{"This node represents an actor",
"This is an action film actor",
"The actor specializes in action",
"This is a comedy film actor",
"The actor specializes in comedy",
"This is a drama film actor",
"The actor specializes in drama"\} \\ 
("movie", "to", "director")
&
\{"The movie is directed by the director",
"The film features direction by the director", 
"The movie's direction was in the hands of the director", 
"The movie was helmed by the director", 
"The film is a directorial effort by the director", 
"The movie bears the directorial signature of the director"\} \\ 
("movie", "to", "actor")
&
\{"The movie has the actor", 
                           "The movie features the actor", 
                           "The film includes the actor in its lineup", 
                           "The movie showcases the talent of the actor", 
                           "The film's cast includes the actor", 
                           "The movie presents the actor on its roster"\} \\ 
("director", "to", "movie")
&
\{"This director is responsible for the film's direction", 
                              "The director take the helm for the movie", 
                              "The director steers the production of the movie", 
                              "The director in question crafts the narrative of the film", 
                              "The director provides the creative direction for the film", 
                              "The director orchestrates the making of the movie"\} \\ 
("actor", "to", "movie")
&
\{"The actor appears in the movie", 
                           "The actor is part of the movie's cast", 
                           "The actor stars in the movie", 
                           "The actor is featured in the film", 
                           "The actor has a role in the movie"\} \\
    \\
    \midrule
    \multicolumn{2}{c}{(b) DBLP} \\
    \cmidrule(rl){1-2}
    "paper"
&
\{"This node represents a paper", 
              "A paper in the area of Database", 
              "A paper in the area of Data Mining", 
              "A paper in the area of AI", 
              "A paper in the area of Information Retrieval", 
              "A paper published in a conference", 
              "A paper in the area of computer science"\} \\
    "conference" & \{"This node represents a conference", 
                   "A conference in the area of Database", 
                   "A conference in the area of Data Mining", 
                   "A conference in the area of AI", 
                   "A conference in the area of Information Retrieval", 
                   "A conference about computer science"\} \\
    "author" & \{"This node represents an author", 
               "An author in the area of Database", 
               "An author in the area of Data Mining", 
               "An author in the area of AI", 
               "An author in the area of Information Retrieval", 
               "An author in the area of computer science"\} \\
    "term" & \{"This node represents a key term related to a paper", 
             "The term is included in a paper", 
             "The term is related to Database", 
             "The term is related to Data Mining", 
             "The term is related to AI",
              "The term is related to Information Retrieval"\} \\
    ("author", "to", "paper") & \{"The author has the paper", 
                            "The author publishes the paper", 
                            "The author writes the paper", 
                            "The author is the author of the paper", 
                            "The author releases the paper", 
                            "The author issues the paper", 
                            "The author disseminates the paper"\} \\ 
("paper", "to", "author") & \{"The paper has the author", 
                            "The paper is published by the author", 
                            "The paper is written by the author",
                            "The paper lists the author", 
                            "The paper is put forth by the author", 
                            "The paper is made public by the author"\} \\ 
("paper", "to", "term") & \{"The paper has the term", 
                          "The paper includes the term",
                          "The paper contains the term", 
                          "The paper encompasses the term", 
                          "The paper lists the term"\} \\ 
("paper", "to", "conference") & \{"The paper is published in the conference", 
                                "The paper is accepted by the conference",
                                "The paper is included in the proceedings of the conference", 
                                "The paper is presented at the conference", 
                                "The paper appears in the conference proceedings", 
                                "The paper is part of the conference's official record.", 
                                "The paper makes its debut at the conference", 
                                "The paper is documented in the conference's scholarly collection"\} \\ 
("term", "to", "paper") & \{"The term is related to the paper", 
                          "The term is included in the paper",
                          "The term is in the paper", 
                          "The term is featured in the paper", 
                          "The term is incorporated into the paper", 
                          "The term is part of the paper's content", 
                          "The term is found within the paper", 
                          "The term appears in the paper"\} \\ 
("conference", "to", "paper") & \{"The conference has the paper", 
                                "The conference includes the paper", 
                                "The conference accepts this paper", 
                                "The proceedings of the conference includes the paper", 
                                "The conference features the publication of the paper", 
                                "The conference includes the paper in its publications", 
                                "The conference serves as the platform for the paper's publication"\} \\

    \bottomrule
  \end{tabular}
  }
\end{table*}

\begin{table*}
  \centering
  \caption{Detailed Descriptions of Text-Enriched Heterogeneity Representations for ACM (Continued).}
  \vspace{-0.15in}
  \label{tab:des_type_2}
  \resizebox{1\linewidth}{!}{
  \begin{tabular}{p{3.3cm}p{15cm}}
    \toprule
    Node (Edge) Type & Sets of Descriptions  \\
    \midrule
    \multicolumn{2}{c}{(c) ACM} \\
    \cmidrule(rl){1-2}
    "paper" & \{"This node represents a paper", 
              "A paper in the area of Database", 
              "A paper in the area of Wireless Communication", 
              "A paper in the area of Data Mining", 
              "A paper published in a conference (one of KDD, SIGMOD, SIGCOMM, MobiCOMM, and VLDB)", 
              "A paper in the area of computer science"\} \\
    "subject" & \{"This node represents a subject", 
                "The subject is related to Database", 
                "The subject is related to Wireless Communication", 
                "The subject is related to Data Mining",
                "The subject is related to computer science"\} \\
    "author" & \{"This node represents an author", 
               "An author in the area of Database", 
               "An author in the area of Data Mining", 
               "An author in the area of AI", 
               "An author in the area of Information Retrieval", 
               "An author in the area of computer science"\} \\
    "term" & \{"This node represents a key term related to a paper", 
             "The term is included in a paper", 
             "The term is related to Database", 
             "The term is related to Wireless Communication", 
             "The term is related to Data Mining",
             "The term is related to computer science"\} \\
    ("author", "to", "paper") & \{"The author has the paper", 
                            "The author publishes the paper", 
                            "The author writes the paper", 
                            "The author is the author of the paper", 
                            "The author releases the paper", 
                            "The author issues the paper", 
                            "The author disseminates the paper"\} \\ 
("paper", "to", "author") & \{"The paper has the author", 
                            "The paper is published by the author", 
                            "The paper is written by the author",
                            "The paper lists the author", 
                            "The paper is put forth by the author", 
                            "The paper is made public by the author"\} \\ 
("paper", "to", "term") & \{"The paper has the term", 
                          "The paper includes the term",
                          "The paper contains the term", 
                          "The paper encompasses the term", 
                          "The paper lists the term"\} \\ 
("term", "to", "paper") & \{"The term is related to the paper", 
                          "The term is included in the paper",
                          "The term is in the paper", 
                          "The term is featured in the paper", 
                          "The term is incorporated into the paper", 
                          "The term is part of the paper's content", 
                          "The term is found within the paper", 
                          "The term appears in the paper"\} \\ 
("paper", "to", "subject") & \{"The paper pertains to the subject", 
                             "The paper is concerned with the subject", 
                             "The paper addresses this subject's matter", 
                             "The paper contributes to the discourse on the subject", 
                             "The paper explores the subject in depth", 
                             "The paper examines the subject", 
                             "The paper is dedicated to the analysis of the subject", 
                             "The paper's content is relevant to the subject", 
                             "The paper provides insights into the subject", 
                             "The paper discusses the subject comprehensively"\} \\ 
("subject", "to", "paper") & \{"The subject serves as the focus for the paper", 
                             "The subject is the central theme of the paper", 
                             "The subject forms the basis of the paper's inquiry", 
                             "The subject underpins the scholarly work presented in the paper", 
                             "The subject informs the paper's research focus", 
                             "The subject delineates the scope of the paper's investigation", 
                             "The subject is the cornerstone of the paper's theoretical foundation"\} \\ 
("paper", "cite", "paper") & \{"The paper cites the paper", 
                             "The paper includes references to a previous paper", 
                             "The manuscript references earlier research in its bibliography", 
                             "The study attributes findings to an antecedent scholarly paper", 
                             "The paper assimilates insights from a previously published study", \} \\ 
("paper", "ref", "paper") & \{"The paper is cited by the paper", 
                            "The paper receives a citation from a subsequent publication", 
                            "The article is referenced within the bibliography of another scholarly work", 
                            "The manuscript is acknowledged by another study in its references", 
                            "The document is listed in the citations of another academic article"\} \\ 
    \\
    \bottomrule
  \end{tabular}}
\end{table*}

\begin{table*}
  \centering
  \caption{Detailed Prompt of Mixture-of-Thought (MoT) Graph Instruction Augmentation.}
  \vspace{-0.15in}
  \label{tab:mot_prompt_app}
  \resizebox{1\linewidth}{!}{
  \begin{tabular}{p{3cm}p{11cm}p{4cm}}
    \toprule
    Prompt Engineering & Prompting Template & Augmented Instruction Template  \\
    \midrule
    \cmidrule(rl){1-3}
    \textbf{CoT without Format Constraint} & I have a question as below: \{Human Question\} ; and the answer is \{Ground Truth\}, imagine that you have made the correct choice and proceed with step-by-step reasoning. Example: Data mining. Based on ... &
     \{Human Question\} $\longrightarrow$ \{GPT's Answer\&Reasoning\} \\ \midrule
     \textbf{CoT with Format Constraint} & I have a question as below: \{Human Question\} ; and the answer is \{Ground Truth\}, imagine that you have made the correct choice and proceed with step-by-step reasoning. Using the following format: Answer: [Answer] Reason: ... &
     \{Human Question\} $\longrightarrow$ \{GPT's Answer\&Reasoning\} \\ \midrule
     \textbf{ToT with Multiple Round} & I have a question as below: \{Human Question\} ; and the answer is \{Ground Truth\}, imagine three different experts are answering this question. All experts will write down 1 step of their thinking, then share it with the group. Then all experts will go on to the next step, etc. If any expert realises they're wrong at any point then they leave. And finally they make the correct choice. Using the following format: Expert 1:... Expert 2:... Expert 3:... Expert 1:... Expert 2:... Expert 3:... Final Answer:... &
     \{Human Question\} $\longrightarrow$ \{GPT's Answer\&Reasoning\} \\ \midrule
     \textbf{ToT with Single Round} & I have a question as below: \{Human Question\} ; and the answer is \{Ground Truth\}, imagine three different experts are answering this question. All experts will write down 1 step of their thinking, then share it with the group. Then all experts will go on to the next step, etc. If any expert realises they're wrong at any point then they leave. And finally they make the correct choice. Using the following format: Expert 1:... Expert 2:... Expert 3:... Final Answer:... &
     \{Human Question\} $\longrightarrow$ \{GPT's Answer\&Reasoning\} \\ \midrule
    \textbf{PanelGPT} & I have a question as below: \{Human Question\} ; and the answer is \{Ground Truth\}, imagine that 3 experts are discussing the question with a panel discussion, trying to solve it step by step to make sure the result is correct and avoid penalty. And finally they make the correct choice. &
     \{Human Question\} $\longrightarrow$ \{GPT's Answer\&Reasoning\} \\ \midrule
     \textbf{GKP-1} & I have a question as below: \{Human Question\} ; and the answer is \{Ground Truth\}, please generate some knowledge that can assist in formulating an answer, including, but not limited to: distinctions between the four categories. Imagine that you have arrived at the correct answer based on the provided information and knowledge, and present a step-by-step reasoning. Using the following format: Knowledge: ... Answer: ... Reason: ... &
     \{Human Question\} +\{GPT's Knowledge\}  $\longrightarrow$ \{GPT's Answer\&Reasoning\} \\ \midrule
    \textbf{GKP-2} & Please generate some knowledge that can assist in formulating an answer, including, but not limited to: explanations of some technical terms present in the given information. Imagine that you have arrived at the correct answer based on the provided information and knowledge, and present a step-by-step reasoning. Using the following format: Knowledge: ... Answer: ... Reason: ... &
     \{Human Question\} +\{GPT's Knowledge\}  $\longrightarrow$ \{GPT's Answer\&Reasoning\} \\
    \\
    \bottomrule
  \end{tabular}}
\end{table*}
\subsection{Instruction and Prompting Templates}\label{sec:inst_ds}
\subsubsection{\bf Detailed prompt of MoT graph instruction augmentation}
In Table~\ref{tab:mot_prompt_app}, we display all MoT graph instruction augmentation strategies that incorporate various prompt engineering techniques, including the prompting template to prompt GPT-3.5, and the template for constructing Instructions after obtaining the results.

\subsubsection{\bf Instruction construction template for two stage instruction tuning}
In Table~\ref{tab:inst_temp_2_app}, we showcase the instruction template for the second stage of node classification for IMDB, where the prompting suffix allows our instruction to be combined with a variety of different prompt techniques for instruction tuning, thereby distilling multiple reasoning abilities for the powerful GPT-3.5. Table~\ref{tab:inst_temp_1_app} presents two types of instruction templates for the first stage, targeting heterogeneous relation awareness and homogeneous relation awareness, respectively.
The construction method for other datasets is the same as that for IMDB.
\subsubsection{\bf Instruction construction template for graph in-context learning}
\begin{figure}[h]
    \centering
    \includegraphics[width=0.5\textwidth]{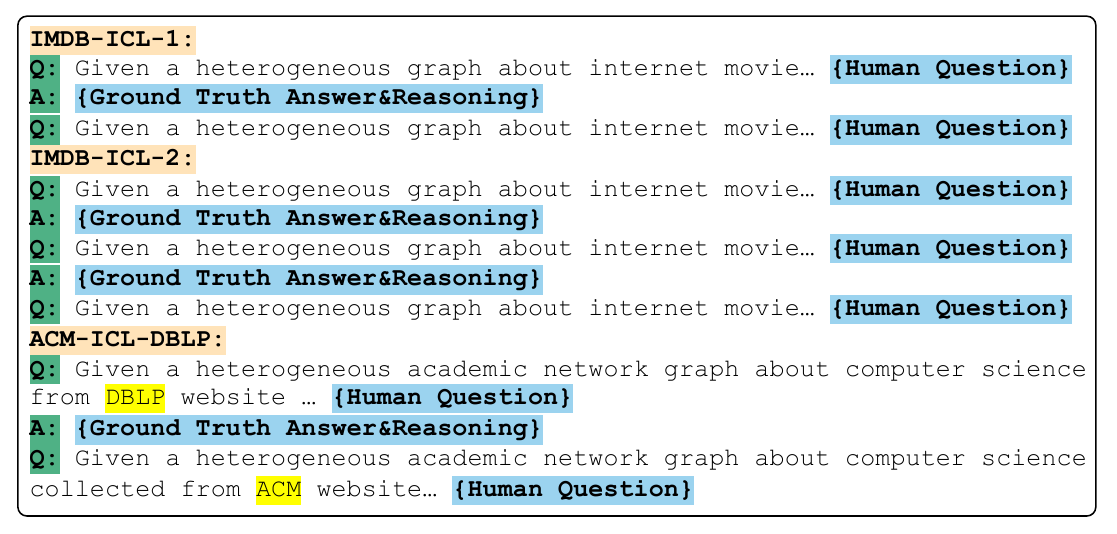}
    \vspace{-0.15in}
    \caption{Instruction construction template for graph in-context learning, including "ICL-1", "ICL-2" and "ICL-DBLP".}
    \vspace{-0.15in}
    \label{fig:graph_icl_inst}
\end{figure}

In Figure~\ref{fig:graph_icl_inst}, we illustrate the construction of instructions for our graph in-context learning tests, which includes "-ICL-1" with one example, "-ICL-2" with two examples, and "-ICL-DBLP" where examples from ACM are concatenated with DBLP examples. We represent the examples and the final question using a \texttt{Q:...A:...Q:...} sequence.
\begin{table*}[h]
\vspace{-0.05in}
\centering
\caption{Instruction template for IMDB.}
\vspace{-0.1in}
    \label{tab:inst_temp_2_app}
\small

\begin{tabularx}{1.02\textwidth}{X X}
    \toprule
     
      \multicolumn{2}{@{}p{1.02\textwidth}@{}}{\centering \textbf{Node Classification}} \\
      \midrule
      \textbf{Input:} Given a heterogeneous graph about internet movie, there are three types of nodes, namely: movie, actor, director. The relationships (meta paths) between different nodes include: [movie is directed by director], [movie has actor]. By performing random sampling of 2-hop 10 neighbors centered on the target movie node, a heterogeneous subgraph is obtained. In the subgraph, "movie" nodes: \texttt{<graph>}, where the 0-th node is the central node that represents a moive with the following information: Name: \texttt{\{movie name\}} Director's name: \texttt{\{director name\}} Actors' names: \texttt{\{actor name\}} Plot keywords: \texttt{\{plot keywords\}} "actor" nodes: \texttt{<graph>}; "director" nodes: \texttt{<graph>}. Question: Which of the following classes does this movie belong to: action, comedy, drama? \texttt{\{Prompting Suffix\}}\\
      \midrule
      \textbf{Output:} \texttt{\{Answer (\& Reasoning)\}}. \\
      \midrule
      \multicolumn{2}{@{}p{1.02\textwidth}@{}}{\centering \textbf{Prompting Suffix of Different Prompting Techniques}} \\
      \midrule
      \textbf{Standard with the direct answer:} Give likely categories directly.\\
      \textbf{CoT without Format Constraint:} Please think about the categorization in a step by step manner and avoid making false associations. Then provide your reasoning.\\
      \textbf{CoT with Format Constraint:} Please think about the categorization in a step by step manner and avoid making false associations. Then provide your reasoning. Using the following format: Answer: [The answer] Reason: ... \\
      \textbf{ToT:} Imagine three different experts are answering this question. All experts will write down 1 step of their thinking, then share it with the group. Then all experts will go on to the next step, etc. If any expert realises they're wrong at any point then they leave. \\
      \textbf{PanelGPT:} 3 experts are discussing the question with a panel discussion, trying to solve it step by step, and make sure the result is correct and avoid penalty. \\
     
     \bottomrule
     
\end{tabularx}
\vspace{-0.1in}
\end{table*}

\begin{table*}[h]
\centering
\caption{Instruction template for IMDB (Continued).}\label{tab:inst_temp_1_app}
\vspace{-0.1in}
    \label{tab:inst_temp_2}
\small

\begin{tabularx}{1.02\textwidth}{X X}
    \toprule
     
      \multicolumn{2}{@{}p{1.02\textwidth}@{}}{\centering \textbf{Instruction for Heterogeneous Relation Awareness}} \\
      \midrule
      \textbf{Input:} Given a heterogeneous graph about internet movie, there are three types of nodes, namely: movie, actor, director. The relationships (meta paths) between different nodes include: [movie is directed by director], [movie has actor]. By performing random sampling of 2-hop 10 neighbors centered on the target movie node, a heterogeneous subgraph is obtained. In the subgraph, there are several sequences of heterogeneous graph nodes of different types as follows:1. \texttt{<graph>} 2. \texttt{<graph>} 3. \texttt{<graph>} Question: please sequentially provide the categories for the node sequences. Example: 1. actor 2. movie 3. director.\\
      \midrule
      \textbf{Output:} \texttt{\{The correct sequence of node types\}}. \\
      \midrule
      \multicolumn{2}{@{}p{1.02\textwidth}@{}}{\centering \textbf{Instruction for Homogeneous Relation Awareness}} \\
      \midrule
      Given a heterogeneous graph about internet movie, there are three types of nodes, namely: movie, actor, director. The relationships (meta paths) between different nodes include: [movie is directed by director], [movie has actor]. By performing random sampling of 2-hop 10 neighbors centered on the target movie node, a heterogeneous subgraph is obtained. In the subgraph, there are "movie" nodes: <graph>. Here is a list of movies' names, their directors' names, their actors' names and their plot keywords: \texttt{\{movie lists\}} Question: please reorder the list of movies according to the order of graph tokens (\ie, complete the matching of graph tokens and movies)\\
      \midrule
      \textbf{Output:} \texttt{\{The correct sequence of movies' names\}}. \\
     
     \bottomrule
     
\end{tabularx}
\vspace{-0.05in}
\end{table*}

\subsection{Supplementary Experimental Results}\label{sec:exp_app}
\subsubsection{\bf Implementation Details}\label{sec:impl}
In the implementation of our \model, we employ heterogeneous graph data with corresponding textual contents from IMDB and DBLP to conduct text-graph contrastive alignment and obtain a heterogeneous graph tokenizer. In the first phase, we utilize a heterogeneous graph corpus consisting of IMDB and DBLP for instruction tuning. We set the batch size to 1 per GPU and train for 1 epoch with the learning rate $2e^{-5}$, the warmup ratio $3e^{-2}$ and the weight decay $1e^{-4}$. The projector obtained from the first phase training is used as the initial state for the second phase, where we set the epochs to 15 and further perform instruction tuning on downstream tasks. The base model used in both stages is vicuna-7B-v1.5, with the maximum context length set to 2048. And  
For the evaluation of most baselines, we utilize their publicly available code.
We employ a grid-search strategy based on default hyperparameter settings to ensure a fair evaluation. For further details, please refer to our released source code.


\subsubsection{\bf Model Case Study}
\label{sec:case_study}
In this subsection, we explore the behavior of our \model\ under different prompting techniques. Specifically, we utilize various prompting techniques to prompt the 10-shot IMDB model, obtaining six different responses, and the prediction cases for different categories of \model\ are shown in Tables~\ref{tab:case_1},~\ref{tab:case_2}, and~\ref{tab:case_3}, respectively. The parts showing the final answers are highlighted in \sethlcolor{pink}\hl{pink}. We make the following observations: \textbf{Obs.1} Our \model, after instruction tuning with the MoT graph instruction augmentation strategy, can dynamically respond accurately to different prompts. \textbf{Obs.2} The CoT prompt in Table~\ref{tab:case_1}, which is unformatted, also shows a certain format (highlighted in \sethlcolor{yellow}\hl{yellow}), which is attributed to the fact that mixing a variety of instructions can also benefit different prompting techniques. \textbf{Obs.3} As highlighted in \sethlcolor{green}\hl{green} in multiple cases, our \model, after our designed two-stage graph instruction tuning, is consciously considering issues from a graph perspective, further proving that our model is not only solving downstream problems from a textual perspective but has also developed a certain level of graph-awareness.

\begin{table*}[h]
\vspace{-0.00in}
\centering
\caption{Visualization of our \model's response with different prompting engineering techniques on IMDB for action genre.}
\vspace{-0.1in}
    \label{tab:case_1}
\small
\begin{tabularx}{1.02\textwidth}{X X}
    \toprule
     
    \textbf{Question:} Given a heterogeneous graph about internet movie, there are three types of nodes, namely: movie, actor, director. The relationships (meta paths) between different nodes include: [movie is directed by director], [movie has actor]. By performing random sampling of 2-hop 10 neighbors centered on the target movie node, a heterogeneous subgraph is obtained. In the subgraph, "movie" nodes: <graph>, where the 0-th node is the central node that represents a movie with the following information: Name: Live Free or Die Hard; Director's name: Len Wiseman; Actors' names: Bruce Willis|Jonathan Sadowski|Cyril Raffaelli; Plot keywords: fbi|hacker|independence day|police|terrorist; "actor" nodes: <graph>; "director" nodes: <graph>. Question: Which of the following classes does this movie belong to: action, comedy, drama? \\
      
      \midrule
      \textbf{Ground-Truth Category:} Action \\
      \midrule
      \textcolor{violet}{+ Give likely categories directly.} \textbf{\model\ (Direct Response):} Action. \\
      \midrule
      \textcolor{violet}{+ Please think about the categorization in a step by step manner and avoid making false associations. Then provide your reasoning.} \\ \textbf{\model\ (CoT Response without Format Constraint):} \sethlcolor{yellow}\hl{Answer: Action Reasoning:} The movie "Live Free or Die Hard" has several plot keywords that suggest an action genre, such as "fbi", "hacker", "independence day", "police", and "terrorist". These keywords imply that the movie is likely to have a lot of action scenes, chases, and possibly some thrilling moments. Additionally, the director Len Wiseman has directed other action movies in the past, such as "Underworld" and "Total Recall". This further supports the idea that "Live Free or Die Hard" belongs to the action genre. Finally, the actors in the movie also contribute to the action genre. Bruce Willis, who plays the lead role, is known for his roles in action movies such as "Die Hard", "Pulp Fiction", and "The Fifth Element". Jonathan Sadowski and Cyril Raffaelli, who also appear in the movie, have acted in other action movies and TV shows. Overall, based on the plot keywords, the director's past work, and the actors' roles, it is reasonable to conclude that "Live Free or Die Hard" belongs to the action genre.
       \\
        \midrule
        \textcolor{violet}{+ Please think about the categorization in a step by step manner and avoid making false associations. Then provide your reasoning. Using the following format: Answer: [The answer] Reason: ...}
      \\ \textbf{\model\ (CoT Response with Format Constraint):} \sethlcolor{yellow}\hl{Answer: Action Reason:} Based on the plot keywords provided, the movie seems to involve elements of action, such as the FBI, hackers, police, and terrorists. These keywords suggest that the movie may have a high level of action and suspense, which is a common characteristic of action movies. Additionally, the presence of Bruce Willis, a well-known action star, further supports the classification of this movie as an action film. \\
      \midrule
        \textcolor{violet}{+ Imagine three different experts are answering this question. All experts will write down 1 step of their thinking, then share it with the group. Then all experts will go on to the next step, etc. If any expert realizes they're wrong at any point then they leave.}
      \\ \textbf{\model\ (ToT Response):} Expert 1: First, I will analyze the plot keywords of the movie. The keywords "fbi", "hacker", "independence day", "police", and "terrorist" suggest that the movie might belong to the action genre. However, the presence of the word "independence day" could also indicate that the movie might be a combination of action and drama. Expert 2: Next, I will look at the director's name. Len Wiseman is known for directing the "Underworld" film series, which are primarily action movies. Therefore, it is likely that "Live Free or Die Hard" belongs to the \sethlcolor{pink}\hl{action genre.} Expert 3:Finally, I will examine the actors' names. Bruce Willis is an action star, and Jonathan Sadowski and Cyril Raffaelli are relatively unknown actors. Therefore, it is more likely that the movie belongs to the action genre. \sethlcolor{pink}\hl{Based on the analysis of the plot keywords, the director's name, and the actors' names, it is likely that "Live Free or Die Hard" belongs to the action genre.} \\
      \midrule
        \textcolor{violet}{+ 3 experts are discussing the question with a panel discussion, trying to solve it step by step, and make sure the result is correct and avoid penalty}
      \\ \textbf{\model\ (Panel Discussion Response):} Expert 1: Let's start by analyzing the plot keywords. The keywords "fbi", "hacker", "independence day", "police", and "terrorist" suggest that this movie might belong to the action genre. Expert 2: That's a good point. However, we should also consider the director's previous works. Len Wiseman is known for directing the "Underworld" series, which are primarily action movies. Therefore, it's likely that this movie also belongs to the action genre. Expert 3: I agree with both of you. However, we should also consider the actors' names. Bruce Willis is known for his roles in action movies, and Jonathan Sadowski and Cyril Raffaelli are relatively unknown actors. \sethlcolor{pink}\hl{Therefore, it's possible that this movie might also belong to the drama genre.} Expert 1: That's a good point. We should consider all possible genres. \sethlcolor{green}\hl{Let's perform a random sampling of 2-hop 10 neighbors centered on the target movie node to obtain a heterogeneous subgraph.} Expert 2: Great idea. Let's see what we can find. Expert 3: Agreed. \sethlcolor{green}\hl{After performing the random sampling, the experts found that the movie "Live Free or Die Hard" belongs to the action genre.} \\

     \bottomrule
\end{tabularx}
\vspace{-0.1in}
\end{table*}

\begin{table*}[h]
\vspace{-0.3in}
\centering
\caption{Visualization of our \model's response with different prompting engineering techniques on IMDB for comedy genre.}
\vspace{-0.1in}
    \label{tab:case_2}
\small
\begin{tabularx}{1.02\textwidth}{X X}
    \toprule
     
    \textbf{Question:} Given a heterogeneous graph about internet movie, there are three types of nodes, namely: movie, actor, director. The relationships (meta paths) between different nodes include: [movie is directed by director], [movie has actor]. By performing random sampling of 2-hop 10 neighbors centered on the target movie node, a heterogeneous subgraph is obtained. In the subgraph, "movie" nodes: <graph>, where the 0-th node is the central node that represents a movie with the following information: Name: Caddyshack; Director's name: Harold Ramis; Actors' names: Bill Murray|Rodney Dangerfield|Brian Doyle-Murray; Plot keywords: caddy|country club|golf course|gopher|nudity; "actor" nodes: <graph>; "director" nodes: <graph>. Question: Which of the following classes does this movie belong to: action, comedy, drama? \\
      
      \midrule
      \textbf{Ground-Truth Category:} Comedy \\
      \midrule
      \textcolor{violet}{+ Give likely categories directly.} \textbf{\model\ (Direct Response):} Comedy. \\
      \midrule
      \textcolor{violet}{+ Please think about the categorization in a step by step manner and avoid making false associations. Then provide your reasoning.} \\ \textbf{\model\ (CoT Response without Format Constraint):} Based on the information provided in the subgraph, the movie "Caddyshack" belongs to the comedy genre. The plot keywords "caddy|country club|golf course|gopher|nudity" suggest that the movie is a comedy that involves a golf course, caddy, and potentially some nudity. Additionally, the director Harold Ramis is known for his work in comedy films, such as "Groundhog Day" and "Ghostbusters." To further support the classification of "Caddyshack" as a comedy, \sethlcolor{green}\hl{we can compare it to other movies in the same subgraph} that are also directed by Harold Ramis. For example, "Groundhog Day" is also a comedy, and it shares some of the same actors, such as Bill Murray, with "Caddyshack." Therefore, it is reasonable to assume that "Caddyshack" also belongs to the comedy genre. \sethlcolor{pink}\hl{In summary, the movie "Caddyshack" belongs to the comedy genre} based on its plot keywords, director Harold Ramis' previous work in comedy films, \sethlcolor{green}\hl{and its similarity to other comedies in the same subgraph.}
       \\
        \midrule
        \textcolor{violet}{+ Please think about the categorization in a step by step manner and avoid making false associations. Then provide your reasoning. Using the following format: Answer: [The answer] Reason: ...}
      \\ \textbf{\model\ (CoT Response with Format Constraint):} \sethlcolor{yellow}\hl{Answer: Comedy Reason:} The movie "Caddyshack" is a comedy film, as indicated by its plot keywords such as "caddy", "country club", "golf course", and "gopher", which suggest a light-hearted and humorous tone. Additionally, the director Harold Ramis has a history of directing comedies, further supporting the classification of this movie as a comedy. \\
      \midrule
        \textcolor{violet}{+ Imagine three different experts are answering this question. All experts will write down 1 step of their thinking, then share it with the group. Then all experts will go on to the next step, etc. If any expert realizes they're wrong at any point then they leave.}
      \\ \textbf{\model\ (ToT Response):} Based on the plot keywords, "caddy," "country club," "golf course," and "nudity," it seems like the movie "Caddyshack" belongs to the comedy genre. Expert 1: The movie has keywords like "caddy," "country club," and "golf course," which are commonly associated with comedy movies. Expert 2: The presence of Bill Murray, Rodney Dangerfield, and Brian Doyle-Murray, who are known for their comedic roles, also suggests that the movie is a comedy. Expert 3: The plot keywords "nudity" may be a red herring, as it is not a common characteristic of comedy movies. \sethlcolor{pink}\hl{Therefore, the movie likely belongs to the comedy genre.} \\
      \midrule
        \textcolor{violet}{+ 3 experts are discussing the question with a panel discussion, trying to solve it step by step, and make sure the result is correct and avoid penalty}
      \\ \textbf{\model\ (Panel Discussion Response):} To solve this problem, we need to analyze the plot keywords and the director's name to determine the genre of the movie. The plot keywords suggest that the movie is likely to be a comedy, as it contains keywords such as "caddy," "country club," "golf course," and "gopher," which are commonly associated with comedic situations in movies. The director's name, Harold Ramis, also supports this conclusion. He is known for directing comedies such as "Groundhog Day" and "Ghostbusters," which were both critical and commercial successes. Therefore, we can conclude that the movie "Caddyshack" belongs to the comedy genre. As for the sorting of the "actor" nodes, we can use the plot keywords to determine the order. The actors' names suggest that the movie has a mix of well-known and lesser-known actors, but the plot keywords suggest that the movie is primarily focused on the actors' interactions with the golf course and gopher. Therefore, we can sort the actors' names based on their prominence in the movie, with Bill Murray and Rodney Dangerfield being more well-known and therefore higher up in the list, and Brian Doyle-Murray being lesser-known but still important to the plot. So the sorted list of actors' names would be: 1. Bill Murray 2. Rodney Dangerfield 3. Brian Doyle-Murray As for the sorting of the "movie" nodes, we can use the plot keywords to determine the order. The plot keywords suggest that the movie is primarily focused on the golf course and gopher, so we can sort the movie nodes based on their prominence in the movie, with "Caddyshack" being the central node and the other two movies being lesser-known and therefore lower down in the list. So the sorted list of movie nodes would be: 1. Caddyshack 2. Happy Gilmore 3. The Greatest Game Ever Played Therefore, the final sorted list of nodes would be: 1. Caddyshack 2. Harold Ramis 3. Bill Murray 4. Rodney Dangerfield 5. Brian Doyle-Murray 6. Happy Gilmore 7. The Greatest Game Ever Played Note that the order of the "actor" nodes and the "movie" nodes may vary depending on the specific criteria used to sort them. \sethlcolor{pink}\hl{However, the overall conclusion that "Caddyshack" belongs to the comedy genre} and is \sethlcolor{green}\hl{the central node in the subgraph remains the same.} \\

     \bottomrule
\end{tabularx}
\vspace{-0.1in}
\end{table*}

\begin{table*}[h]
\vspace{-0.3in}
\centering
\caption{Visualization of our \model's response with different prompting engineering techniques on IMDB for drama genre.}
\vspace{-0.1in}
    \label{tab:case_3}
\small
\begin{tabularx}{1.02\textwidth}{X X}
    \toprule
     
    \textbf{Question:} Given a heterogeneous graph about internet movie, there are three types of nodes, namely: movie, actor, director. The relationships (meta paths) between different nodes include: [movie is directed by director], [movie has actor]. By performing random sampling of 2-hop 10 neighbors centered on the target movie node, a heterogeneous subgraph is obtained. In the subgraph, "movie" nodes: <graph>, where the 0-th node is the central node that represents a movie with the following information: Name: The Life of David Gale; Director's name: Alan Parker; Actors' names: Kevin Spacey|Kate Winslet|Matt Craven; Plot keywords: activist|death|death row|murder|texas"actor" nodes: <graph>; "director" nodes: <graph>. Question: Which of the following classes does this movie belong to: action, comedy, drama? \\
      
      \midrule
      \textbf{Ground-Truth Category:} Drama \\
      \midrule
      \textcolor{violet}{+ Give likely categories directly.} \textbf{\model\ (Direct Response):} Drama. \\
      \midrule
      \textcolor{violet}{+ Please think about the categorization in a step by step manner and avoid making false associations. Then provide your reasoning.} \\ \textbf{\model\ (CoT Response without Format Constraint):} Based on the plot keywords provided, the movie "The Life of David Gale" belongs to the drama genre. The plot keywords include "activist," "death," "death row," "murder," and "texas," which suggest a serious and emotional storyline. These keywords are commonly associated with drama films, rather than action or comedy films. \sethlcolor{pink}\hl{Therefore, the movie "The Life of David Gale" belongs to the drama genre.}
       \\
        \midrule
        \textcolor{violet}{+ Please think about the categorization in a step by step manner and avoid making false associations. Then provide your reasoning. Using the following format: Answer: [The answer] Reason: ...}
      \\ \textbf{\model\ (CoT Response with Format Constraint):} \sethlcolor{yellow}\hl{Answer: Drama Reason:} The movie "The Life of David Gale" is a drama, as it deals with serious and emotional themes such as activism, death row, and murder. The plot keywords also suggest a dramatic tone. \\
      \midrule
        \textcolor{violet}{+ Imagine three different experts are answering this question. All experts will write down 1 step of their thinking, then share it with the group. Then all experts will go on to the next step, etc. If any expert realizes they're wrong at any point then they leave.}
      \\ \textbf{\model\ (ToT Response):} Expert 1: Based on the plot keywords, this movie seems to be a drama. Expert 2: I agree with Expert 1. The plot keywords such as "activist," "death," "death row," and "murder" suggest a serious and emotional theme. Expert 3: I also agree with Expert 1 and 2. The plot keywords and the fact that the movie is directed by Alan Parker, who is known for his dramatic films, further support the classification of this movie as a drama. \sethlcolor{pink}\hl{Final answer: The movie belongs to the drama class.} \\
      \midrule
        \textcolor{violet}{+ 3 experts are discussing the question with a panel discussion, trying to solve it step by step, and make sure the result is correct and avoid penalty}
      \\ \textbf{\model\ (Panel Discussion Response):} 1. Expert 1: Let's start by analyzing the plot keywords. We have "activist," "death," "death row," "murder," and "texas." Based on these keywords, I would say this movie belongs to the drama genre. 2. Expert 2: I agree with Expert 1. The plot keywords suggest a serious and emotional storyline. However, we also have "actor" nodes in the graph, which could indicate that there are comedic moments or performances. 3. Expert 3: I would like to add that we should also consider the director's name, Alan Parker. He has directed both dramatic and comedic movies, such as "The Commitments" and "Birdy." Therefore, it's difficult to determine the genre solely based on the plot keywords and director's name. 4. Panel Discussion: Based on the analysis so far, \sethlcolor{pink}\hl{we can conclude that the movie "The Life of David Gale" belongs to the drama genre}, but it may \sethlcolor{green}\hl{also have comedic elements due to the presence of actor nodes in the graph.} \\

     \bottomrule
\end{tabularx}
\vspace{-0.1in}
\end{table*}

\subsubsection{\bf Comprehensive Results of Overall Performance Comparison}
Table~\ref{tab:performance_app} showcases the results of our \model\ in both few-shot and zero-shot settings, covering scenarios with 1, 3, 5, 10, 20, 40, and 60 shots. The results clearly indicate that our model consistently outperforms state-of-the-art models in all cases.
\begin{table*}
\centering
\caption{Comprehensive performance comparison on node classification tasks in both few-shot and zero-shot settings. However, since SSL methods focus on learning embeddings from downstream graphs, we excluded the zero-shot settings for them ("-").}
\label{tab:performance_app}
\resizebox{1\textwidth}{!}{\begin{tabular}{c|c|c|c|clcccccc|cc} 
\hline
Datasets                     & Metric                 & train-on & test-on   & sage           & \multicolumn{1}{c}{gat} & han           & hgt           & hgnn           & dmgi          & HGMAE         & HeCo          & \textbf{ours-std}      & \textbf{ours-cot}       \\ 
\hline
\multirow{21}{*}{Supervised} & \multirow{7}{*}{Mi-F1} & IMDB-1   & IMDB-1000 & 0.4663$\pm$0.0025  & 0.4567$\pm$0.0122           & 0.4890$\pm$0.0271 & 0.4977$\pm$0.0186 & 0.4790$\pm$0.0134  & 0.4570$\pm$0.0126 & 0.3609$\pm$0.0145 & 0.3874$\pm$0.0159 & 0.5090$\pm$0.0073          & \textbf{0.5360$\pm$0.0065}  \\
                             &                        & IMDB-3   & IMDB-1000 & 0.5083$\pm$0.0103  & 0.4970$\pm$0.0100           & 0.4860$\pm$0.0122 & 0.5007$\pm$0.0163 & 0.5027$\pm$0.0066  & 0.4120$\pm$0.0153 & 0.3791$\pm$0.0060 & 0.3467$\pm$0.0208 & 0.5170$\pm$0.0093          & \textbf{0.5730$\pm$0.0108}  \\
                             &                        & IMDB-5   & IMDB-1000 & 0.5010$\pm$0.0051  & 0.5170$\pm$0.0029           & 0.4840$\pm$0.0094 & 0.5003$\pm$0.0093 & 0.5020$\pm$0.0045  & 0.4413$\pm$0.0173 & 0.3652$\pm$0.0062 & 0.3385$\pm$0.0169 & 0.6180$\pm$0.0027          & \textbf{0.6320$\pm$0.0085}  \\
                             &                        & IMDB-10  & IMDB-1000 & 0.5203$\pm$0.0005  & 0.5227$\pm$0.0085           & 0.4967$\pm$0.0111 & 0.5273$\pm$0.0060 & 0.5127$\pm$0.0087  & 0.4710$\pm$0.0148 & 0.3705$\pm$0.0071 & 0.3572$\pm$0.0159 & 0.5710$\pm$0.0092          & \textbf{0.5810$\pm$0.0064}  \\
                             &                        & IMDB-20  & IMDB-1000 & 0.5930$\pm$0.0093  & 0.6117$\pm$0.0012           & 0.5763$\pm$0.0046 & 0.5750$\pm$0.0065 & 0.5957$\pm$0.0054  & 0.5497$\pm$0.0256 & 0.4107$\pm$0.0106 & 0.3781$\pm$0.0148 & 0.6090$\pm$0.0255          & \textbf{0.6440$\pm$0.0075}  \\
                             &                        & IMDB-40  & IMDB-1000 & 0.6170$\pm$0.0112  & 0.6261$\pm$0.0015           & 0.6198$\pm$0.0025 & 0.5923$\pm$0.0040 & 0.6177$\pm$0.0046  & 0.5813$\pm$0.0033 & 0.3946$\pm$0.0067 & 0.3927$\pm$0.0134 & 0.6260$\pm$0.0057          & \textbf{0.6280$\pm$0.0071}  \\
                             &                        & IMDB-60  & IMDB-1000 & 0.6285$\pm$0.0077  & 0.6299$\pm$0.0027           & 0.6311$\pm$0.0057 & 0.6037$\pm$0.0085 & 0.6267$\pm$0.0107  & 0.5920$\pm$0.0086 & 0.4192$\pm$0.0126 & 0.4051$\pm$0.0120 & \textbf{0.6350$\pm$0.0074} & 0.5980$\pm$0.0110           \\ 
\cline{2-14}
                             & \multirow{7}{*}{Ma-F1} & IMDB-1   & IMDB-1000 & 0.4425$\pm$0.0068  & 0.3974$\pm$0.0183           & 0.4229$\pm$0.0104 & 0.4020$\pm$0.0112 & 0.4456$\pm$0.0036  & 0.4083$\pm$0.0288 & 0.3573$\pm$0.0117 & 0.4023$\pm$0.0137 & 0.4986$\pm$0.0141          & \textbf{0.5247$\pm$0.0061}  \\
                             &                        & IMDB-3   & IMDB-1000 & 0.4737$\pm$0.0207  & 0.4388$\pm$0.0057           & 0.4365$\pm$0.0265 & 0.4590$\pm$0.0417 & 0.4784$\pm$0.0208  & 0.4070$\pm$0.0102 & 0.3708$\pm$0.0050 & 0.3647$\pm$0.0232 & 0.5079$\pm$0.0082          & \textbf{0.5591$\pm$0.0070}  \\
                             &                        & IMDB-5   & IMDB-1000 & 0.4613$\pm$0.0086  & 0.4767$\pm$0.0098           & 0.4695$\pm$0.0037 & 0.4676$\pm$0.0153 & 0.4677$\pm$0.0145  & 0.4254$\pm$0.0124 & 0.3500$\pm$0.0080 & 0.3468$\pm$0.0213 & 0.6111$\pm$0.0091          & \textbf{0.6243$\pm$0.0060}  \\
                             &                        & IMDB-10  & IMDB-1000 & 0.5001$\pm$0.0087  & 0.5145$\pm$0.0058           & 0.4971$\pm$0.0103 & 0.5188$\pm$0.0085 & 0.5024$\pm$0.0045  & 0.4699$\pm$0.0125 & 0.3668$\pm$0.0064 & 0.3768$\pm$0.0195 & 0.5694$\pm$0.0162          & \textbf{0.5762$\pm$0.0108}  \\
                             &                        & IMDB-20  & IMDB-1000 & 0.5953$\pm$0.0095  & 0.6121$\pm$0.0024           & 0.5756$\pm$0.0051 & 0.5723$\pm$0.0056 & 0.5969$\pm$0.0055  & 0.5495$\pm$0.0270 & 0.4065$\pm$0.0089 & 0.3904$\pm$0.0172 & 0.6068$\pm$0.0146          & \textbf{0.6398$\pm$0.0083}  \\
                             &                        & IMDB-40  & IMDB-1000 & 0.6182$\pm$0.0107  & 0.6254$\pm$0.0009           & 0.6224$\pm$0.0057 & 0.5909$\pm$0.0068 & 0.6234$\pm$0.0038  & 0.5786$\pm$0.0064 & 0.3866$\pm$0.0072 & 0.3988$\pm$0.0147 & \textbf{0.6265$\pm$0.0090} & 0.6237$\pm$0.0059           \\
                             &                        & IMDB-60  & IMDB-1000 & 0.62913$\pm$0.0062 & 0.6288$\pm$0.0019           & 0.6330$\pm$0.0056 & 0.5999$\pm$0.0109 & 0.6303$\pm$0.0097  & 0.5894$\pm$0.0078 & 0.4017$\pm$0.0114 & 0.4093$\pm$0.0125 & \textbf{0.6328$\pm$0.0058} & 0.5925$\pm$0.0071           \\ 
\cline{2-14}
                             & \multirow{7}{*}{AUC}   & IMDB-1   & IMDB-1000 & 0.6079$\pm$0.0061  & 0.6151$\pm$0.0065           & 0.6234$\pm$0.0252 & 0.6249$\pm$0.0170 & 0.6107$\pm$0.0075  & 0.5780$\pm$0.0130 & 0.5274$\pm$0.0058 & 0.5712$\pm$0.0099 & 0.6565$\pm$0.0146          & \textbf{0.6685$\pm$0.0037}  \\
                             &                        & IMDB-3   & IMDB-1000 & 0.6271$\pm$0.0075  & 0.6301$\pm$0.0066           & 0.6176$\pm$0.0099 & 0.6170$\pm$0.0166 & 0.6275$\pm$0.0047  & 0.5679$\pm$0.0116 & 0.5507$\pm$0.0052 & 0.5347$\pm$0.0290 & 0.6648$\pm$0.0032          & \textbf{0.6935$\pm$0.0098}  \\
                             &                        & IMDB-5   & IMDB-1000 & 0.6309$\pm$0.0049  & 0.6372$\pm$0.0012           & 0.6102$\pm$0.0059 & 0.6197$\pm$0.0152 & 0.6290$\pm$0.0022  & 0.5832$\pm$0.0132 & 0.5262$\pm$0.0041 & 0.5067$\pm$0.0228 & 0.7308$\pm$0.0125          & \textbf{0.7310$\pm$0.0086}  \\
                             &                        & IMDB-10  & IMDB-1000 & 0.6401$\pm$0.0015  & 0.6435$\pm$0.0062           & 0.6254$\pm$0.0064 & 0.6411$\pm$0.0036 & 0.6360$\pm$0.0051  & 0.6108$\pm$0.0088 & 0.5310$\pm$0.0068 & 0.5432$\pm$0.0189 & \textbf{0.6968$\pm$0.0065} & 0.6875$\pm$0.0083           \\
                             &                        & IMDB-20  & IMDB-1000 & 0.6976$\pm$0.0064  & 0.7122$\pm$0.0020           & 0.6815$\pm$0.0052 & 0.6801$\pm$0.0048 & 0.7005$\pm$0.0030  & 0.6657$\pm$0.0179 & 0.5766$\pm$0.0064 & 0.5541$\pm$0.0145 & 0.7227$\pm$0.0034          & \textbf{0.7424$\pm$0.0113}  \\
                             &                        & IMDB-40  & IMDB-1000 & 0.7171$\pm$0.0069  & 0.7210$\pm$0.0014           & 0.7204$\pm$0.0015 & 0.6970$\pm$0.0060 & 0.7145$\pm$0.0035  & 0.6860$\pm$0.0027 & 0.5488$\pm$0.0049 & 0.5653$\pm$0.0105 & 0.7323$\pm$0.0036          & \textbf{0.7331$\pm$0.0074}  \\
                             &                        & IMDB-60  & IMDB-1000 & 0.7298$\pm$0.0064  & 0.7262$\pm$0.0017           & 0.7281$\pm$0.0013 & 0.6998$\pm$0.0141 & 0.7191$\pm$0.0079  & 0.6934$\pm$0.0063 & 0.5725$\pm$0.0066 & 0.5809$\pm$0.0124 & \textbf{0.7371$\pm$0.0053} & 0.7120$\pm$0.0141           \\ 
\hline
\multirow{42}{*}{Zero-shot}  & \multirow{7}{*}{Mi-F1} & IMDB-1   & DBLP-1000 & 0.2353$\pm$0.0372  & 0.1893$\pm$0.0373           & 0.2653$\pm$0.0203 & 0.2573$\pm$0.0519 & 0.2900$\pm$0.0638  & ~-~           & ~-~           & ~-~           & 0.3180$\pm$0.0072          & \textbf{0.3500$\pm$0.0073}  \\
                             &                        & IMDB-3   & DBLP-1000 & 0.2627$\pm$0.0085  & 0.2913$\pm$0.0017           & 0.2510$\pm$0.0000 & 0.2143$\pm$0.0569 & 0.2430$\pm$0.0453  & ~-~           & ~-~           & ~-~           & 0.3150$\pm$0.0108          & \textbf{0.3660$\pm$0.0060}  \\
                             &                        & IMDB-5   & DBLP-1000 & 0.2607$\pm$0.0082  & 0.2737$\pm$0.0176           & 0.2577$\pm$0.0094 & 0.2453$\pm$0.0458 & 0.2427$\pm$0.0452  & ~-~           & ~-~           & ~-~           & 0.3180$\pm$0.0044          & \textbf{0.3620$\pm$0.0047}  \\
                             &                        & IMDB-10  & DBLP-1000 & 0.2603$\pm$0.0078  & 0.2957$\pm$0.0105           & 0.2863$\pm$0.0108 & 0.2797$\pm$0.0217 & 0.2417$\pm$0.0441  & ~-~           & ~-~           & ~-~           & 0.3490$\pm$0.0029          & \textbf{0.4020$\pm$0.0108}  \\
                             &                        & IMDB-20  & DBLP-1000 & 0.2810$\pm$0.0289  & 0.2780$\pm$0.0033           & 0.2710$\pm$0.0000 & 0.2803$\pm$0.0208 & 0.2333$\pm$0.0353  & ~-~           & ~-~           & ~-~           & 0.3840$\pm$0.0088          & \textbf{0.4180$\pm$0.0083}  \\
                             &                        & IMDB-40  & DBLP-1000 & 0.2400$\pm$0.0324  & 0.2847$\pm$0.0053           & 0.2710$\pm$0.0000 & 0.2937$\pm$0.0005 & 0.2027$\pm$0.0345  & ~-~           & ~-~           & ~-~           & 0.3320$\pm$0.0087          & \textbf{0.3630$\pm$0.0045}  \\
                             &                        & IMDB-60  & DBLP-1000 & 0.2350$\pm$0.0377  & 0.2887$\pm$0.0123           & 0.2710$\pm$0.0000 & 0.2937$\pm$0.0005 & 0.2170$\pm$0.0274  & ~-~           & ~-~           & ~-~           & 0.3520$\pm$0.0032          & \textbf{0.3950$\pm$0.0104}  \\ 
\cline{2-14}
                             & \multirow{7}{*}{Ma-F1} & IMDB-1   & DBLP-1000 & 0.0963$\pm$0.0132  & 0.1169$\pm$0.0089           & 0.1047$\pm$0.0063 & 0.1016$\pm$0.0169 & 0.1778$\pm$0.0629  & ~-~           & ~-~           & ~-~           & 0.2048$\pm$0.0068          & \textbf{0.2472$\pm$0.0070}  \\
                             &                        & IMDB-3   & DBLP-1000 & 0.1099$\pm$0.0094  & 0.1295$\pm$0.0121           & 0.1003$\pm$0.0000 & 0.0939$\pm$0.0149 & 0.0972$\pm$0.0148  & ~-~           & ~-~           & ~-~           & 0.2011$\pm$0.0106          & \textbf{0.2814$\pm$0.0071}  \\
                             &                        & IMDB-5   & DBLP-1000 & 0.1042$\pm$0.0028  & 0.1291$\pm$0.0145           & 0.1024$\pm$0.0030 & 0.1138$\pm$0.0296 & 0.0971$\pm$0.0148  & ~-~           & ~-~           & ~-~           & 0.1917$\pm$0.0046          & \textbf{0.2773$\pm$0.0085}  \\
                             &                        & IMDB-10  & DBLP-1000 & 0.1033$\pm$0.0024  & 0.1420$\pm$0.0155           & 0.1113$\pm$0.0033 & 0.1330$\pm$0.0165 & 0.0969$\pm$0.0145  & ~-~           & ~-~           & ~-~           & 0.2455$\pm$0.0091          & \textbf{0.3386$\pm$0.0093}  \\
                             &                        & IMDB-20  & DBLP-1000 & 0.1448$\pm$0.0573  & 0.1274$\pm$0.0060           & 0.1066$\pm$0.0000 & 0.1143$\pm$0.0116 & 0.1008$\pm$0.0191  & ~-~           & ~-~           & ~-~           & 0.3142$\pm$0.0074          & \textbf{0.3733$\pm$0.0051}  \\
                             &                        & IMDB-40  & DBLP-1000 & 0.1068$\pm$0.0060  & 0.1588$\pm$0.0078           & 0.1066$\pm$0.0000 & 0.1268$\pm$0.0105 & 0.0984$\pm$0.0161  & ~-~           & ~-~           & ~-~           & 0.2331$\pm$0.0069          & \textbf{0.2912$\pm$0.0056}  \\
                             &                        & IMDB-60  & DBLP-1000 & 0.1013$\pm$0.0040  & 0.1821$\pm$0.0134           & 0.1066$\pm$0.0000 & 0.1164$\pm$0.0040 & 0.1078$\pm$0.0282  & ~-~           & ~-~           & ~-~           & 0.2620$\pm$0.0048          & \textbf{0.3359$\pm$0.0114}  \\ 
\cline{2-14}
                             & \multirow{7}{*}{AUC}   & IMDB-1   & DBLP-1000 & 0.4999$\pm$0.0001  & 0.4513$\pm$0.0295           & 0.5000$\pm$0.0000 & 0.5000$\pm$0.0000 & 0.5206$\pm$0.0306  & ~-~           & ~-~           & ~-~           & 0.5222$\pm$0.0069          & \textbf{0.5406$\pm$0.0040}  \\
                             &                        & IMDB-3   & DBLP-1000 & 0.4997$\pm$0.0005  & 0.4995$\pm$0.0016           & 0.5000$\pm$0.0000 & 0.4933$\pm$0.0095 & 0.5000$\pm$0.0000  & ~-~           & ~-~           & ~-~           & 0.5185$\pm$0.0134          & \textbf{0.5524$\pm$0.0073}  \\
                             &                        & IMDB-5   & DBLP-1000 & 0.4978$\pm$0.0030  & 0.4908$\pm$0.0078           & 0.5000$\pm$0.0000 & 0.5031$\pm$0.0043 & 0.4998$\pm$0.0003  & ~-~           & ~-~           & ~-~           & 0.5184$\pm$0.0081          & \textbf{0.5493$\pm$0.0091}  \\
                             &                        & IMDB-10  & DBLP-1000 & 0.4975$\pm$0.0030  & 0.5022$\pm$0.0068           & 0.5000$\pm$0.0000 & 0.5018$\pm$0.0058 & 0.4992$\pm$0.0007  & ~-~           & ~-~           & ~-~           & 0.5396$\pm$0.0071          & \textbf{0.5777$\pm$0.0060}  \\
                             &                        & IMDB-20  & DBLP-1000 & 0.5154$\pm$0.0213  & 0.4918$\pm$0.0020           & 0.5000$\pm$0.0000 & 0.5011$\pm$0.0016 & 0.4957$\pm$0.0060  & ~-~           & ~-~           & ~-~           & 0.5669$\pm$0.0041          & \textbf{0.5907$\pm$0.0089}  \\
                             &                        & IMDB-40  & DBLP-1000 & 0.5027$\pm$0.0031  & 0.4976$\pm$0.0021           & 0.5000$\pm$0.0000 & 0.5008$\pm$0.0006 & 0.4884$\pm$0.0164  & ~-~           & ~-~           & ~-~           & 0.5296$\pm$0.0070          & \textbf{0.5508$\pm$0.0086}  \\
                             &                        & IMDB-60  & DBLP-1000 & 0.4987$\pm$0.0019  & 0.5034$\pm$0.0053           & 0.5000$\pm$0.0000 & 0.5000$\pm$0.0000 & 0.4952$\pm$0.0068  & ~-~           & ~-~           & ~-~           & 0.5426$\pm$0.0071          & \textbf{0.5732$\pm$0.0089}  \\ 
\cline{2-14}
                             & \multirow{7}{*}{Mi-F1} & IMDB-1   & ACM-1000  & 0.3293$\pm$0.0418  & 0.3567$\pm$0.0053           & 0.3407$\pm$0.0111 & 0.3240$\pm$0.0014 & 0.3743$\pm$0.0434  & ~-~           & ~-~           & ~-~           & 0.4160$\pm$0.0106          & \textbf{0.4540$\pm$0.0089}  \\
                             &                        & IMDB-3   & ACM-1000  & 0.3987$\pm$0.0228  & 0.3933$\pm$0.0495           & 0.4400$\pm$0.0177 & 0.3277$\pm$0.0060 & 0.4167$\pm$0.0090  & ~-~           & ~-~           & ~-~           & 0.4530$\pm$0.0086          & \textbf{0.5000$\pm$0.0078}  \\
                             &                        & IMDB-5   & ACM-1000  & 0.3820$\pm$0.0113  & 0.3787$\pm$0.0057           & 0.3630$\pm$0.0086 & 0.3160$\pm$0.0169 & 0.3583$\pm$0.0198  & ~-~           & ~-~           & ~-~           & 0.4580$\pm$0.0173          & \textbf{0.4880$\pm$0.0131}  \\
                             &                        & IMDB-10  & ACM-1000  & 0.3097$\pm$0.0147  & 0.2940$\pm$0.0033           & 0.3073$\pm$0.0026 & 0.3343$\pm$0.0132 & 0.3093$\pm$0.0025  & ~-~           & ~-~           & ~-~           & 0.4850$\pm$0.0078          & \textbf{0.5160$\pm$0.0187}  \\
                             &                        & IMDB-20  & ACM-1000  & 0.2807$\pm$0.0074  & 0.3013$\pm$0.0188           & 0.3133$\pm$0.0031 & 0.3530$\pm$0.0000 & 0.2840$\pm$0.0226  & ~-~           & ~-~           & ~-~           & \textbf{0.5080$\pm$0.0129} & 0.5030$\pm$0.0064           \\
                             &                        & IMDB-40  & ACM-1000  & 0.3173$\pm$0.0005  & 0.2393$\pm$0.0144           & 0.2697$\pm$0.0194 & 0.3560$\pm$0.0099 & 0.3180$\pm$0.0016  & ~-~           & ~-~           & ~-~           & 0.4750$\pm$0.0149          & \textbf{0.5050$\pm$0.0077}  \\
                             &                        & IMDB-60  & ACM-1000  & 0.3400$\pm$0.0057  & 0.2520$\pm$0.0104           & 0.2573$\pm$0.0034 & 0.3593$\pm$0.0037 & 0.3333$\pm$0.0103  & ~-~           & ~-~           & ~-~           & 0.4310$\pm$0.0064          & \textbf{0.5120$\pm$0.0051}  \\ 
\cline{2-14}
                             & \multirow{7}{*}{Ma-F1} & IMDB-1   & ACM-1000  & 0.2647$\pm$0.0269  & 0.2908$\pm$0.0131           & 0.2250$\pm$0.0416 & 0.1631$\pm$0.0005 & 0.3139$\pm$0.0468  & ~-~           & ~-~           & ~-~           & 0.3949$\pm$0.0078          & \textbf{0.4177$\pm$0.0124}  \\
                             &                        & IMDB-3   & ACM-1000  & 0.3131$\pm$0.0393  & 0.2820$\pm$0.0714           & 0.3528$\pm$0.0208 & 0.1710$\pm$0.0114 & 0.3587$\pm$0.0249  & ~-~           & ~-~           & ~-~           & 0.3999$\pm$0.0056          & \textbf{0.4516$\pm$0.0116}  \\
                             &                        & IMDB-5   & ACM-1000  & 0.3208$\pm$0.0130  & 0.3009$\pm$0.0137           & 0.2782$\pm$0.0026 & 0.1969$\pm$0.0301 & 0.3087$\pm$0.0225  & ~-~           & ~-~           & ~-~           & 0.4336$\pm$0.0085          & \textbf{0.4510$\pm$0.0114}  \\
                             &                        & IMDB-10  & ACM-1000  & 0.2497$\pm$0.0065  & 0.2022$\pm$0.0136           & 0.2078$\pm$0.0017 & 0.1670$\pm$0.0049 & 0.2471$\pm$0.0062  & ~-~           & ~-~           & ~-~           & 0.4218$\pm$0.0081          & \textbf{0.4598$\pm$0.0054}  \\
                             &                        & IMDB-20  & ACM-1000  & 0.2694$\pm$0.0091  & 0.2422$\pm$0.0098           & 0.2412$\pm$0.0050 & 0.2094$\pm$0.0501 & 0.2715$\pm$0.0181  & ~-~           & ~-~           & ~-~           & \textbf{0.4964$\pm$0.0075} & 0.4877$\pm$0.0070           \\
                             &                        & IMDB-40  & ACM-1000  & 0.3117$\pm$0.0017  & 0.2141$\pm$0.0071           & 0.2313$\pm$0.0132 & 0.2749$\pm$0.0122 & 0.3144$\pm$0.0017  & ~-~           & ~-~           & ~-~           & 0.4176$\pm$0.0116          & \textbf{0.4585$\pm$0.0089}  \\
                             &                        & IMDB-60  & ACM-1000  & 0.3308$\pm$0.0069  & 0.2141$\pm$0.0071           & 0.2389$\pm$0.0024 & 0.2258$\pm$0.0299 & 0.3226$\pm$0.0074  & ~-~           & ~-~           & ~-~           & 0.3703$\pm$0.0074          & \textbf{0.4862$\pm$0.0083}  \\ 
\cline{2-14}
                             & \multirow{7}{*}{AUC}   & IMDB-1   & ACM-1000  & 0.4934$\pm$0.0247  & 0.5248$\pm$0.0038           & 0.5128$\pm$0.0086 & 0.5000$\pm$0.0000 & 0.5318$\pm$0.0295  & ~-~           & ~-~           & ~-~           & 0.5672$\pm$0.0040          & \textbf{0.5969$\pm$0.0082}  \\
                             &                        & IMDB-3   & ACM-1000  & 0.5559$\pm$0.0173  & 0.5523$\pm$0.0379           & 0.5880$\pm$0.0136 & 0.5028$\pm$0.0040 & 0.5695$\pm$0.0064  & ~-~           & ~-~           & ~-~           & 0.5970$\pm$0.0032          & \textbf{0.6323$\pm$0.0130}  \\
                             &                        & IMDB-5   & ACM-1000  & 0.5433$\pm$0.0082  & 0.5415$\pm$0.0047           & 0.5282$\pm$0.0073 & 0.4950$\pm$0.0134 & 0.5256$\pm$0.0145  & ~-~           & ~-~           & ~-~           & 0.5991$\pm$0.0103          & \textbf{0.6224$\pm$0.0054}  \\
                             &                        & IMDB-10  & ACM-1000  & 0.4860$\pm$0.0116  & 0.4748$\pm$0.0018           & 0.4851$\pm$0.0022 & 0.5000$\pm$0.0000 & 0.4853$\pm$0.0022  & ~-~           & ~-~           & ~-~           & 0.6219$\pm$0.0100          & \textbf{0.6450$\pm$0.0076}  \\
                             &                        & IMDB-20  & ACM-1000  & 0.4601$\pm$0.0048  & 0.4772$\pm$0.0137           & 0.4877$\pm$0.0029 & 0.5038$\pm$0.0053 & 0.4625$\pm$0.0163q & ~-~           & ~-~           & ~-~           & \textbf{0.6352$\pm$0.0094} & 0.6318$\pm$0.0068           \\
                             &                        & IMDB-40  & ACM-1000  & 0.4867$\pm$0.0013  & 0.4320$\pm$0.0108           & 0.4545$\pm$0.0146 & 0.5148$\pm$0.0043 & 0.4872$\pm$0.0006  & ~-~           & ~-~           & ~-~           & 0.6138$\pm$0.0047          & \textbf{0.6360$\pm$0.0051}  \\
                             &                        & IMDB-60  & ACM-1000  & 0.5085$\pm$0.0041  & 0.4439$\pm$0.0078           & 0.4473$\pm$0.0026 & 0.5070$\pm$0.0044 & 0.5037$\pm$0.0082  & ~-~           & ~-~           & ~-~           & 0.5801$\pm$0.0051          & \textbf{0.6399$\pm$0.0058}  \\
\hline
\end{tabular}}
\end{table*}

\subsubsection{\bf Comprehensive Results of Graph In-Context Learning}
\begin{figure*}[t]
  \vspace{-0.05in}
  \centering
    
  \subfigure[\bf IMDB-IMDB@Mi-F1, Ma-F1, AUC]{
      \centering
      \includegraphics[width=0.33\textwidth]{resources/in_context/imdb_mif1_new.pdf}
      \includegraphics[width=0.33\textwidth]{resources/in_context/imdb_maf1_new.pdf}
      \includegraphics[width=0.33\textwidth]{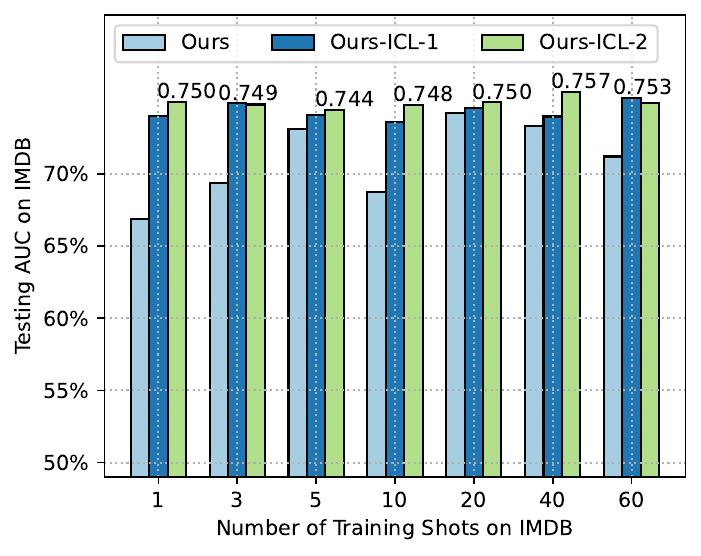}
  }\vspace{-0.15in}
  \subfigure[\bf IMDB-DBLP@Mi-F1, Ma-F1, AUC]{
      \centering
      \includegraphics[width=0.33\textwidth]{resources/in_context/dblp_mif1.pdf}
      \includegraphics[width=0.33\textwidth]{resources/in_context/dblp_maf1.pdf}
      \includegraphics[width=0.33\textwidth]{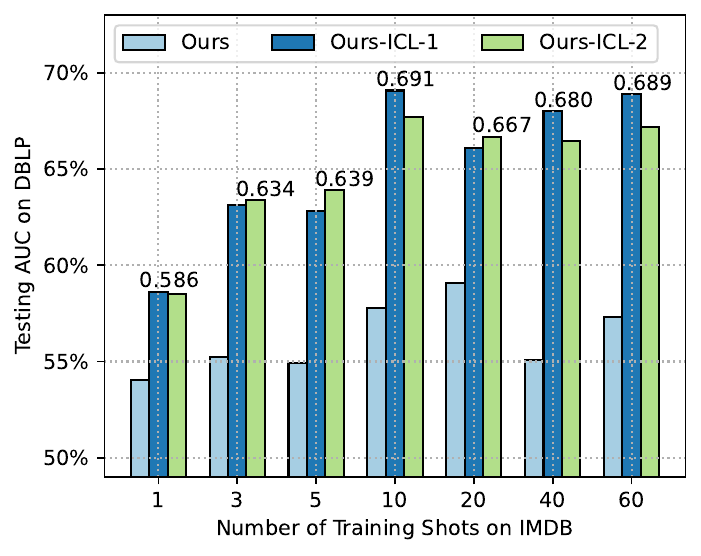}
  }\vspace{-0.15in}
  \subfigure[\bf IMDB-ACM@Mi-F1, Ma-F1, AUC]{
      \centering
      \includegraphics[width=0.33\textwidth]{resources/in_context/acm_mif1.pdf}
      \includegraphics[width=0.33\textwidth]{resources/in_context/acm_maf1.pdf}
      \includegraphics[width=0.33\textwidth]{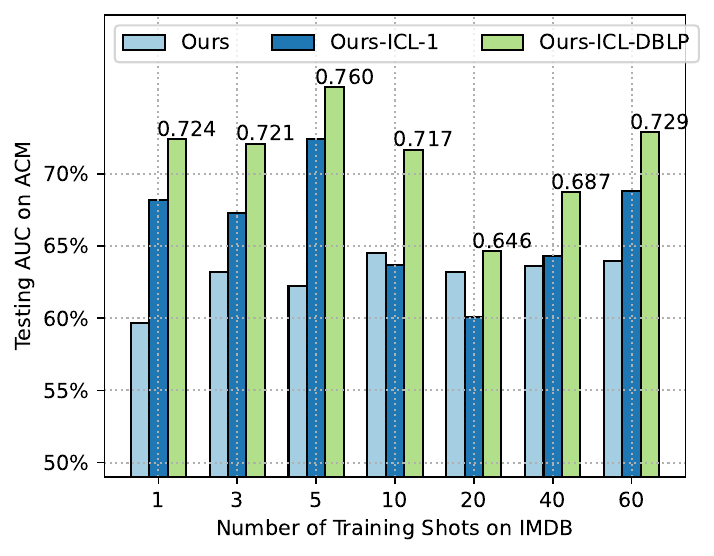}
  }\vspace{-0.2in}
      \caption{Comprehensive results of graph in-context learning of our \model. }\label{fig:graph_icl_app}
  \vspace{-0.2in}

\end{figure*}

The performance of our model on all metrics, across different shot scenarios and datasets, under the graph in-context learning is illustrated in Figure~\ref{fig:graph_icl_app}. The results clearly demonstrate that our graph in-context learning approach significantly improves the model's performance without any modifications to the model parameters.

\end{document}